\documentclass[letterpaper, 10 pt, conference]{ieeeconf}  

\IEEEoverridecommandlockouts                              

\usepackage[english]{babel}
\usepackage{amsmath} 
\usepackage{amssymb}  
\usepackage{graphicx}
\usepackage{caption}
\usepackage{soul}
\usepackage{booktabs}
\usepackage{multirow}
\usepackage{dsfont}
\usepackage{amsfonts}
\usepackage{relsize}
\usepackage{tabularx}
\usepackage{algorithm,algpseudocode}
\usepackage{bm}
\usepackage{xurl}
\usepackage{mdframed}
\usepackage{enumerate}
\usepackage{graphicx}
\usepackage{cite}
\usepackage{xcolor}
\usepackage{float}
\usepackage{siunitx}
\usepackage{mdframed}
\usepackage{authblk}
\usepackage{subcaption}
\makeatletter
\let\NAT@parse\undefined
\makeatother
\usepackage{hyperref}

\title{\LARGE \bf
Long-range UAV Thermal Geo-localization with Satellite Imagery
}

\author{Jiuhong Xiao$^1$, Daniel Tortei$^2$, Eloy Roura$^2$, and Giuseppe Loianno$^1$ 
\thanks{$^1$The authors are with the New York University, Tandon School of Engineering, Brooklyn, NY 11201, USA.  {\tt\footnotesize email: \{jx1190, loiannog\}@nyu.edu.}}
\thanks{$^2$The authors are with the Autonomous Robotics Research Center-Technology Innovation Institute, Abu Dhabi, UAE. {\tt\footnotesize email:  \{daniel.tortei, eloy.roura\}.@tii.ae}.}
\thanks{This work was supported by the Technology Innovation Institute, Qualcomm Research, Nokia, and NYU Wireless. Giuseppe Loianno serves as consultant for the Technology Innovation Institute. This arrangement has been reviewed and approved by the New York University in accordance with its policy on objectivity in research.}
}

\newcommand{\GAN}{{\text{GAN}}}
\newcommand{\TGM}{{\text{TGM}}}
\newcommand{\sat}{{\text{S}}}
\newcommand{\thr}{{\text{T}}}

\begin{document}

\maketitle
\thispagestyle{empty}
\pagestyle{empty}

\begin{abstract}
Onboard sensors, such as cameras and thermal sensors, have emerged as effective alternatives to Global Positioning System (GPS) for geo-localization in Unmanned Aerial Vehicle (UAV) navigation. Since GPS can suffer from signal loss and spoofing problems, researchers have explored camera-based techniques such as Visual Geo-localization (VG) using satellite RGB imagery. Additionally, thermal geo-localization (TG) has become crucial for long-range UAV flights in low-illumination environments. This paper proposes a novel thermal geo-localization framework using satellite RGB imagery, which includes multiple domain adaptation methods to address the limited availability of paired thermal and satellite images. The experimental results demonstrate the effectiveness of the proposed approach in achieving reliable thermal geo-localization performance, even in thermal images with indistinct self-similar features. We evaluate our approach on real data collected onboard a UAV. We also release the code and \textit{Boson-nighttime}, a dataset of paired satellite-thermal and unpaired satellite images for thermal geo-localization with satellite imagery. To the best of our knowledge, this work is the first to propose a thermal geo-localization method using satellite RGB imagery in long-range flights. 
\end{abstract}

\section*{Supplementary Material}

\noindent \textbf{Code and dataset:} \url{https://github.com/arplaboratory/satellite-thermal-geo-localization}
\section{Introduction}~\label{sec:introduction}
Unmanned Aerial Vehicle (UAV) long-range navigation has become an increasingly popular research area due to UAV's ability to address practical issues related to environmental inspection, monitoring, and coverage. Geo-localization is an essential part of the UAV long-range navigation pipeline. Global Positioning System (GPS) is a commonly-used measurement for geo-localization, but it suffers from signal loss and spoofing issues \cite{review_avl}. Using onboard sensors such as cameras due to their low cost, ease of use, and low weight became a popular technique for onboard localization. Visual Inertial Odometry (VIO) is a popular technique for obtaining onboard localization in a GPS-denied or GPS-spoofing environment. Visual measurements can provide precise location information in short-range flights without external signals but have non-negligible drifting errors in long-range navigation settings without loop closure detection \cite{driftingerror}. Similarly, Visual Geo-localization (VG) with satellite RGB imagery is another solution for localization under GPS-denied environments particularly suitable for long-range flight operations since drifting-free. These methods typically try to match UAV images with satellite images in a pre-built database. They require UAVs to fly beyond a certain altitude to capture usable UAV images. In this work, we address the challenging task of VG task in a low-illumination environment during high-altitude long-range UAV flights using thermal imagery. The limited availability of satellite-thermal paired data and the presence of low-contrast self-similar thermal features make this task particularly challenging. The visual inconsistency between thermal and optical features degrades the matching between satellite and thermal images. For example, the thermal camera may not detect an object with distinguishable color for minor infrared radiation.

\begin{figure}
    \centering
    \includegraphics[width=0.45\textwidth]{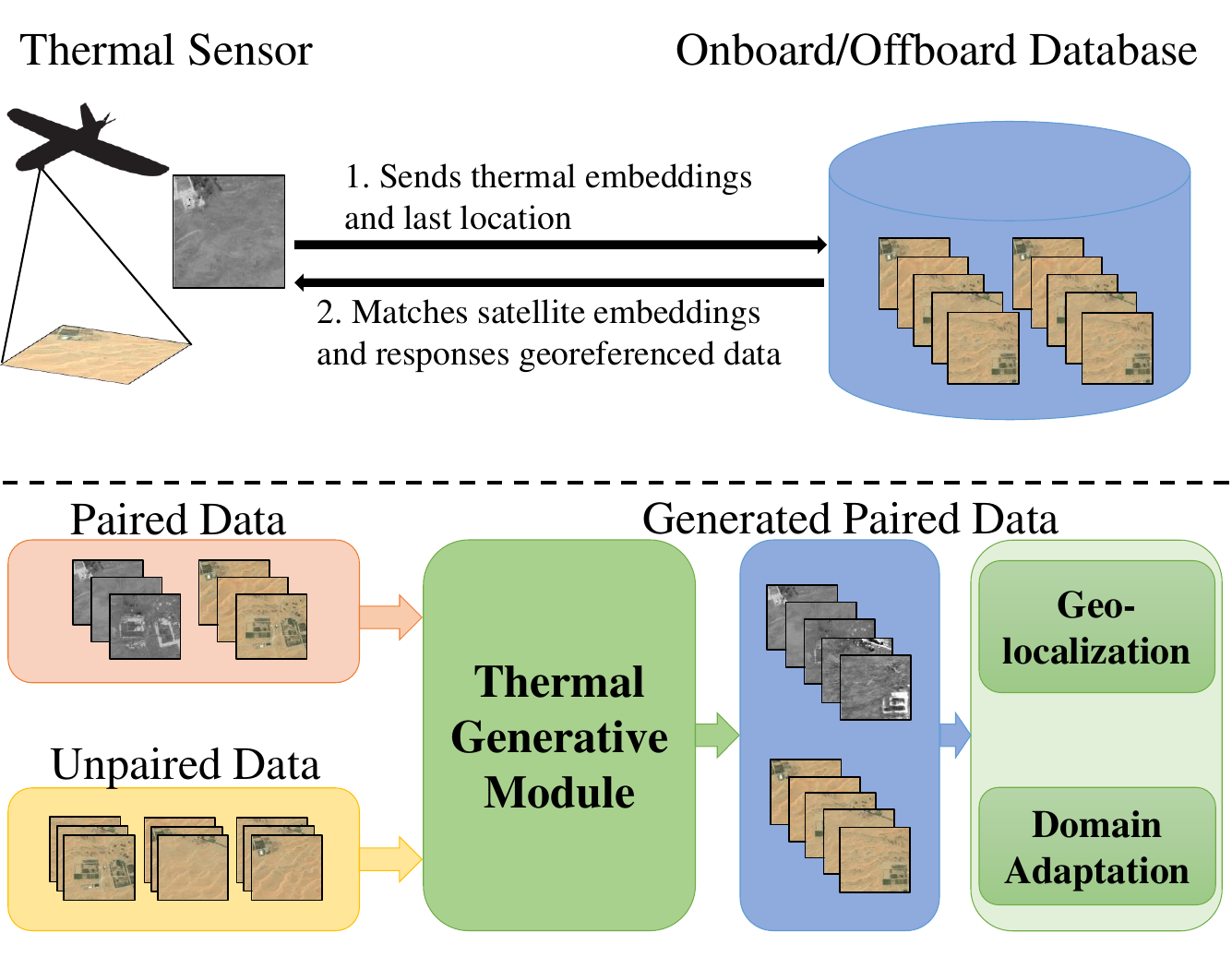}
    \caption{Thermal geo-localization with satellite imagery. Our method aims to perform geo-localization with a thermal sensor attached to a UAV in a low-illumination environment with a satellite image database (upper part). In the training of the geo-localization model (lower part), the thermal generative module is trained by paired satellite-thermal dataset and generates fake thermal images for a large number of unpaired satellite images. The satellite-thermal geo-localization module takes advantage of generated paired dataset and the domain adaptation method to boost performances.}
    \label{teaser}
\end{figure}

VG with satellite imagery requires accurate and robust matching between satellite and UAV images. 
Traditional image matching methods include alignment-based methods \cite{directalign, directalign2, directalign3} and hand-crafted feature point matching \cite{imgregistration, imgregistration2}. 
In image matching, these are vulnerable to appearance changes (e.g., season-variant scenes) and illumination changes (e.g., day-night conditions). 
Deep-learning-based VG methods \cite{Berton_CVPR_2022_benchmark, netvlad, VG1, patch-netvlad, VG3} have gained popularity since they increase the robustness against the aforementioned changes and improve the matching accuracy and generalization to different types of terrains.
These methods commonly use a deep neural network \cite{lecun2015deep} to extract and match features from UAV and satellite images. 
For the aforementioned reasons, our work on TG will mainly follow the deep-learning-based VG workflow.

Thermal Geo-localization (TG) is a critical part of UAV long-range nighttime flight but is rarely explored compared with VG with satellite imagery. In these operating conditions, the camera is unavailable for the low-illumination environment, and the localization algorithm should instead rely on thermal sensors. However, since the thermal sensor detects infrared radiation rather than visible light, the captured images of thermal sensors generally have notable visual inconsistency compared to those obtained using visible spectrum cameras. In these cases, the geo-localization algorithm should be appropriately adapted to operate with satellite-thermal matching.

The main contributions of this work are as follows. First, we propose a novel learning-based thermal geo-localization method using satellite RGB imagery for high-altitude long-range UAV flights (Fig. \ref{teaser}). Second, we show how to use two complementary domain adaptation methods in the proposed approach to compensate for the degraded performance due to the limited availability of thermal data. Third, we present several experimental results on the \textbf{Boson-nighttime} dataset and prove the effectiveness of the proposed approach to guarantee reliable thermal geo-localization performance in the test region with self-similar thermal features. Finally, we release the code and dataset to the community to collaborate and contribute to pushing the boundaries of this research area. To the best of our knowledge, the proposed method is the first geo-localization method using satellite RGB and thermal imagery to achieve TG in long-range flights.
\section{Related Works}~\label{sec:relatedworks}
The goal of TG with satellite imagery is to match high-resolution colored satellite images with grayscale low-resolution thermal images to retrieve the georeference data of matched satellite images. This problem is challenging for low-contrast self-similar thermal features and the significant representation differences between thermal and visible camera data. This section mainly discusses existing visual geo-localization methods with satellite imagery, previous work on UAV thermal localization, and domain adaptation technology to address the domain gap between satellite and thermal data. 

\textbf{Visual Geo-localization.} VG is defined as recognizing the capturing location of given camera images with meter-level error tolerance. Previous works \cite{bow,vlad} mainly follow the image retrieval paradigm. First, the algorithms build a database made of compacted image descriptors from a human-designed feature extractor (e.g., SIFT \cite{SIFT}) using images from search regions. Second, the algorithms use the same feature extractor to generate image descriptors for query images. Finally, they retrieve the image with the highest similarity to the query from the database and get the corresponding geographical information. Recent works \cite{Berton_CVPR_2022_benchmark, netvlad, VG1, patch-netvlad, VG3} investigate the use of deep convolutional neural networks \cite{lecun2015deep} to optimize a learned feature extractor on a large landmark dataset \cite{msls} for better recall performance. Notably, NetVLAD \cite{netvlad} can perform feature aggregation to generate compacted and distinguishable global descriptors from the local feature. Deep-VG-benchmark \cite{Berton_CVPR_2022_benchmark} has recently been released as a framework to benchmark multiple VG methods under a standard setting for datasets, training, and evaluation. Compared to these works, our work stresses the combination of visual geo-localization and domain adaptation technologies to improve thermal geo-localization performance.

VG with satellite imagery is considered an absolute self-localization method \cite{review_avl} in UAV localization research. The satellite map is an important reference data source for localization since it provides top-down images with offset-free longitude and latitude information for a large target region. For traditional image matching methods, direct alignment methods \cite{directalign, directalign2, directalign3} densely match UAV images and satellite images with the highest pixel similarity. Image registration methods \cite{imgregistration, imgregistration2} employ hand-crafted feature point descriptors to match feature points between satellite and UAV images. Recent works \cite{voplusvg, style, VGUAV3, VGUAV4, VGUAV5} also introduce deep neural networks like general VG tasks. Specifically, previous works such as \cite{style} use conditional generative adversarial nets \cite{cgan} to synthesize a UAV-view image with a satellite image style. The synthesis mitigates the domain gap between satellite and UAV images. In \cite{voplusvg}, the authors combine visual odometry and a cross-view geo-localization module to predict the location of UAVs with a Kalman filter. In contrast to these works, our proposed method dedicates to cross-domain geo-localization between satellite RGB and thermal images.

\textbf{UAV Thermal Localization.} Relevant works for UAV thermal localization \cite{thermaluav1, thermaluav2, thermaluav3, thermaluav4} study the localization and navigation performance of Thermal-Inertial Odometry (TIO). UAV thermal stereo odometry is proposed in \cite{thermaluav1} to navigate short-range daytime and nighttime flights. Keyframe-based thermal-inertial odometry is proposed in \cite{thermaluav2} for navigation in dark scenarios. The authors in \cite{thermaluav3} use top-down thermal camera settings to investigate localization performance at various times of the night. A Recent work~\cite{thermaluav4} proposes a data-efficient collaborative decentralized TIO for multi-UAV navigation. These works demonstrate that TIO performs on par with daytime VIO in short-range indoor or outdoor flights. Compared to existing works, our proposed approach focuses on geo-localization by satellite-thermal matching for long-range high-altitude flights.

\begin{figure*}
    \centering
    \smallskip
    \smallskip
    \includegraphics[width=0.9\textwidth]{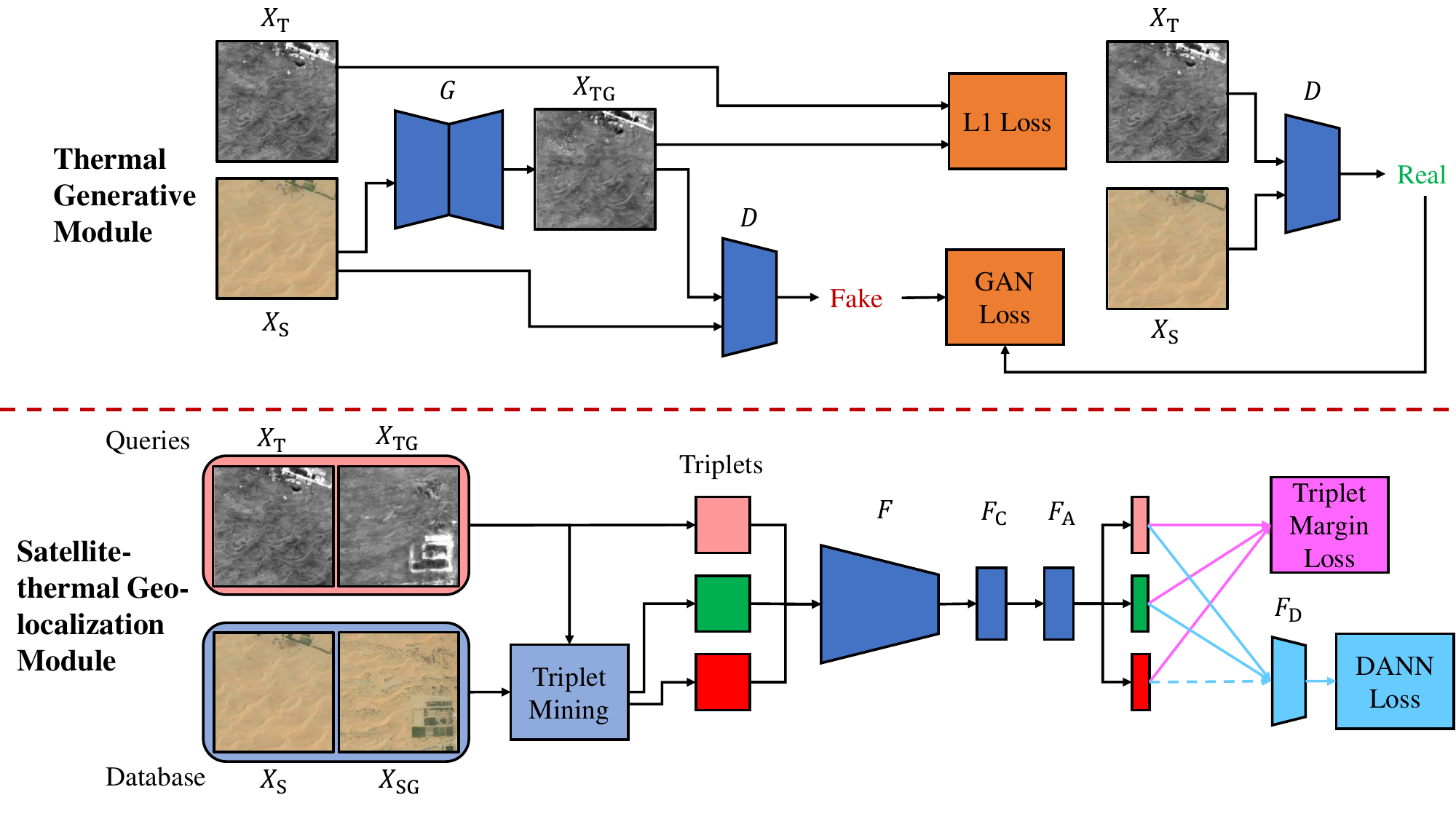}
    \caption{The proposed framework of thermal geo-localization. Thermal Generative Module (TGM) is optimized by L1 Loss and least square GAN Loss. After training, the module generates fake thermal images $X_\textrm{TG}$ from unpaired satellite images $X_\textrm{SG}$. Satellite-thermal Geo-localization Module (SGM) finds a positive sample ({\color{green}green}) and a negative sample ({\color{red}red}) relative to the query sample {(\color{pink} pink}). The triplets input the feature extractor $F$, the compression layer $F_\textrm{C}$, and the aggregation module $F_\textrm{A}$, and the outputs are 1-D embeddings. The Triplet Margin Loss ({\color{magenta}magenta}) and DANN Loss ({\color{cyan}cyan}) are optimized for geo-localization models $F$, $F_\textrm{C}$, and $F_\textrm{A}$. The dashed line for $F_\textrm{D}$ means the negative sample is optional for DANN loss.}
    \label{fig:framework}
\end{figure*}

\textbf{Domain Adaptation.} Due to the limited availability of thermal images from UAVs as well as cost-efficiency, domain adaptation methods can enhance the performance of the TG model. Our application defines the satellite and thermal domains as the source and target domains, respectively. We focus on the adversarial-based methods \cite{DANN, cycada, DA1, thermalDA1, thermalDA2, thermalDA3, thermalDA4}. One of the adversarial-based methods, DANN \cite{DANN}, uses a gradient reversal layer and a domain classifier to achieve unsupervised domain adaptation. These modules classify the domain of the features and make the distribution of features from the source and target domain similar. CyCADA \cite{cycada}, one of the adversarial-based methods with a generative model, applies a CycleGAN-based \cite{CycleGAN2017} framework to generate the image with source-domain content and target-domain style for domain adaptation. In \cite{thermalpix2pix}, the authors propose a sparse GAN method based on pix2pix \cite{pix2pix} to generate thermal images from optical images. In our work, we propose a framework combining DANN for domain adaptation in training Satellite-thermal Geo-localization Module (SGM) and the pix2pix \cite{pix2pix} generative model to generate an extended paired satellite-thermal dataset from numerous unpaired satellite images in Thermal Generative Module (TGM).

\section{Methodology}~\label{sec:methodology}
The proposed TG framework, depicted in Fig. \ref{fig:framework}, has two main components: a Thermal Generative Module (TGM) and a Satellite-thermal Geo-localization Module (SGM). In this section, we describe them in detail.

\subsection{Thermal Generative Module (TGM)}\label{sec:TGM}

For the Thermal Generative Module (TGM), we leverage the pix2pix \cite{pix2pix} model. Let $X_\sat, X_\thr$ denote the normalized satellite images and thermal images with height $H$, width $W$, and the number of channels $C$. $G$ is the generator to generate a fake thermal image $X_\textrm{TG}=G(X_\sat)$. $D$ is the conditional discriminator where the first input is $X_\thr$ or $X_\textrm{TG}$ and the second input is $X_\sat$, and the output predicts if the input thermal image is a fake thermal image. The objectives of the module without the random noise are
 \begin{equation}
\begin{split}
    \min_D L_\GAN(D) &= \frac{1}{2}\mathbb{E}_{X_\thr}[(D(X_\thr) - b)^2] \\
    & +\frac{1}{2}\mathbb{E}_{X_\sat}[(D(G(X_\sat)) - a)^2],
\end{split}
\end{equation}
\begin{align}
    \min_G L_\GAN(G) &= 
    \mathbb{E}_{X_\textrm{S}}[(D(G(X_\sat)) - c)^2],\\
    \min_G L_1(G) &= \mathbb{E}_{X_\sat, X_\thr}[\vert G(X_\sat) - X_\thr \vert],
\end{align}
where $L_\GAN$ is the least square GAN loss \cite{lsgan}, $L_1$ is the L1 loss, $a, b, c$ are the fake label, the real label and the label that $G$ wants $D$ to classify for fake data, respectively. The total loss $L_\TGM$ is the weighted sum of $L_\GAN$ and $L_1$ as
\begin{equation}
    L_\TGM(G,D) = L_\GAN(G) + L_\GAN(D) + \lambda_1L_1(G),
\end{equation}
where $\lambda_1$ is the weight of $L_1$ loss and is set to $100.0$ (original setting in \cite{pix2pix}) or $10.0$. We train the $G$ to minimize $L_\GAN(G)$ and $L_1(G)$, and train the $D$ to minimize $L_\GAN(D)$.

\subsection{Satellite-thermal Geo-localization Module (SGM)}\label{sec:SGM}

 The proposed Satellite-thermal Geo-localization Module (SGM) leverages the image retrieval workflow from deep-VG-benchmark\cite{Berton_CVPR_2022_benchmark}. Additionally, we improve it with a compression layer $F_\textrm{C}$ to control the dimensionality of output descriptors, a module to train simultaneously with the paired dataset and the generated dataset, and a DANN loss \cite{DANN} branch for domain adaption. $X_\textrm{S}, X_\textrm{T}$ are a pair of satellite and thermal images. $X_\textrm{SG}$ denotes satellite images without paired thermal images and $X_\textrm{TG}$ denotes the generated thermal images from $X_\textrm{SG}$ using the trained generator $G$. During training, the framework samples thermal images from the thermal queries dataset. Triplet mining searches for the positive (within a nearby region of queries) and the negative (out of the nearby region of queries) satellite database samples that have the lowest $L_2$ distance to the queries in the feature (embedding) space. The satellite database is built by tiling the satellite map with a certain stride (more details are available in Section~\ref{dataset}). Each triplet consists of one query thermal image, one positive, and one negative satellite image. We input the triplets of $H\times W\times C$ (size and the number of channels) to the feature extractor $F$, and the output is feature maps $H/16\times W/16\times C^\prime$ where $C^\prime$ is the number of channels for the feature map. The compression layer $F_\textrm{C}$ consists of a 2-D convolution layer mapping from $H/16\times W/16\times C^\prime$ to $H/16\times W/16\times C_{\textrm{target}}$ with kernel size = $1$ and a 2-D batch normalization layer. $C_{\textrm{target}}$ is target channel number we want to control. The aggregation module $F_\textrm{A}$ uses NetVLAD \cite{netvlad} to map the compressed feature maps $H/16\times W/16\times C_{\textrm{target}}$ to 1-D embeddings $1\times C_{\textrm{final}}$ where $C_\textrm{final}$ is the final output dimension. Triplet margin loss $L_\textrm{T}$ is used as
 \begin{equation}
     L_\textrm{T}(q,p,n) = (\Vert q - p\Vert_2 - \Vert q - n\Vert_2 + m)^+,
 \end{equation}
 where $q, p, n$ are the embeddings ($1\times C_\textrm{final}$) of query, positive and negative samples. $m$ is a positive scalar margin and is set to 0.1. $L_\textrm{T}$ aims to decrease the $L_2$ embedding distance of $q$ and $p$ and increase that of $q$ and $n$.

 We also introduce DANN loss \cite{DANN} into SGM for domain adaptation. $F_\textrm{D}$ is a domain classifier and the goal is to classify $q$ as the thermal label $d$ and $p, n$ as the satellite label $e$. A gradient reversal layer is attached to the beginning of $F_\textrm{D}$, and the DANN loss is a cross-entropy loss
 \begin{equation}
    \begin{split}
        L_\textrm{DANN}(q,p,n) &= -\sum_{c\in\left\{d, e\right\}}[y_{c,q}\log(o_{c,q})\\ & + y_{c,p}\log(o_{c,p}) + y_{c,n}\log(o_{c,n})],
    \end{split}
\end{equation}
 where $y_{d,q}=1, y_{e,q}=0$ for thermal embeddings, $y_{d,p}=0, y_{e,p}=1, y_{d,n}=0, y_{e,n}=1$ for satellite embeddings, $o_{c,q}, o_{c,p}, o_{c,n}$ are the output probability of $q, p, n$ for the class $c$ using $F_\textrm{D}$. The reversed gradients backpropagate to make the distribution of $q, p, n$ similar. In the experiments, we find that using negative embeddings $n$ for DANN loss affects the performance since DANN loss may conflict with triplet margin loss on $q, n$. DANN loss optimizes to increase the similarity between the distribution of $q, n$ while triplet margin loss increases $\Vert q - n\Vert_2$. We take $n$ as an optional input for $F_\textrm{D}$. The total loss $L_\textrm{SGM}$ is the weighted sum of $L_\textrm{T}$ and $L_\textrm{DANN}$ as
 \begin{equation}
     L_\textrm{SGM} = L_\textrm{T} + \lambda_2 L_\textrm{DANN},
 \end{equation}
 where $\lambda_2$ is the weight of DANN loss and is set to $0.1$.
\section{Experimental Setup}~\label{sec:experiment_setting}
In this section, we introduce the experimental setup of our proposed framework. We discuss the dataset setting (Section \ref{dataset}), the implementation and experimental details (Section \ref{imp}), and the proposed evaluation metrics (Section~\ref{metrics}).

\begin{figure}
    \centering
    \smallskip
    \smallskip
    \rotatebox{90}{\hspace{2em}Satellite}
    \begin{subfigure}[b]{0.15\textwidth}
    \includegraphics[width=\textwidth]{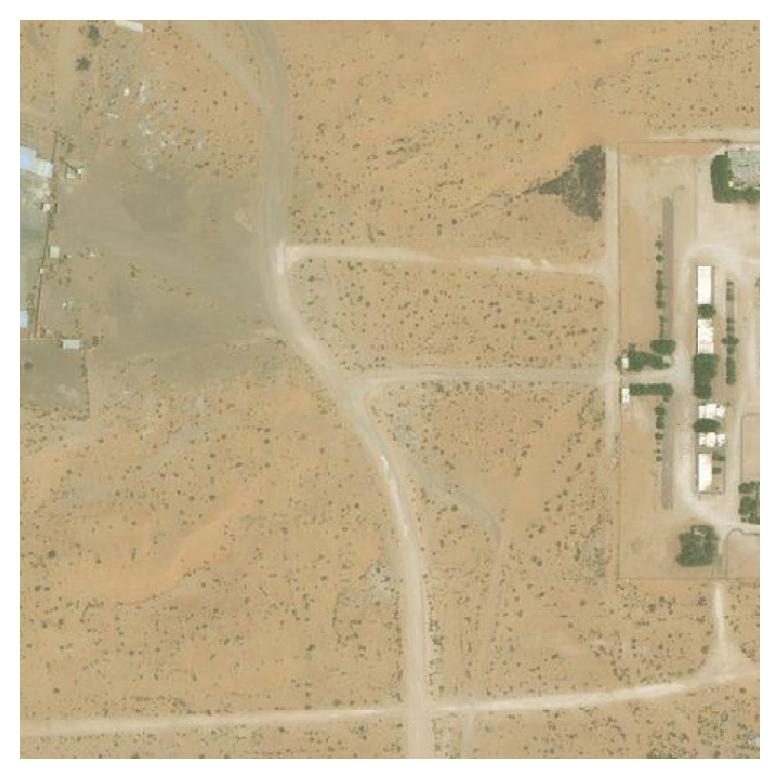}
\end{subfigure}
\begin{subfigure}[b]{0.15\textwidth}
    \includegraphics[width=\textwidth]{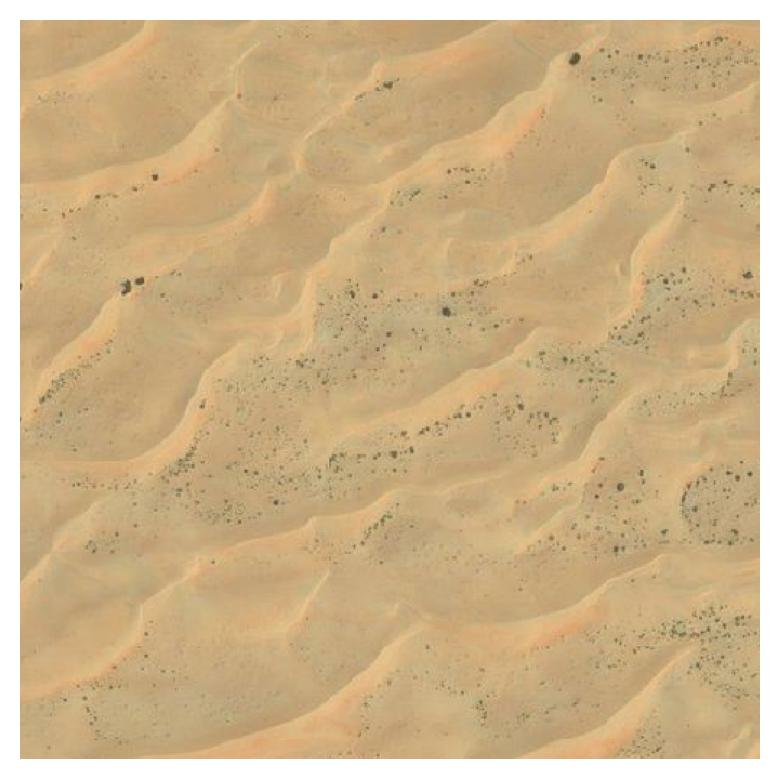}
\end{subfigure}
\begin{subfigure}[b]{0.15\textwidth}
    \includegraphics[width=\textwidth]{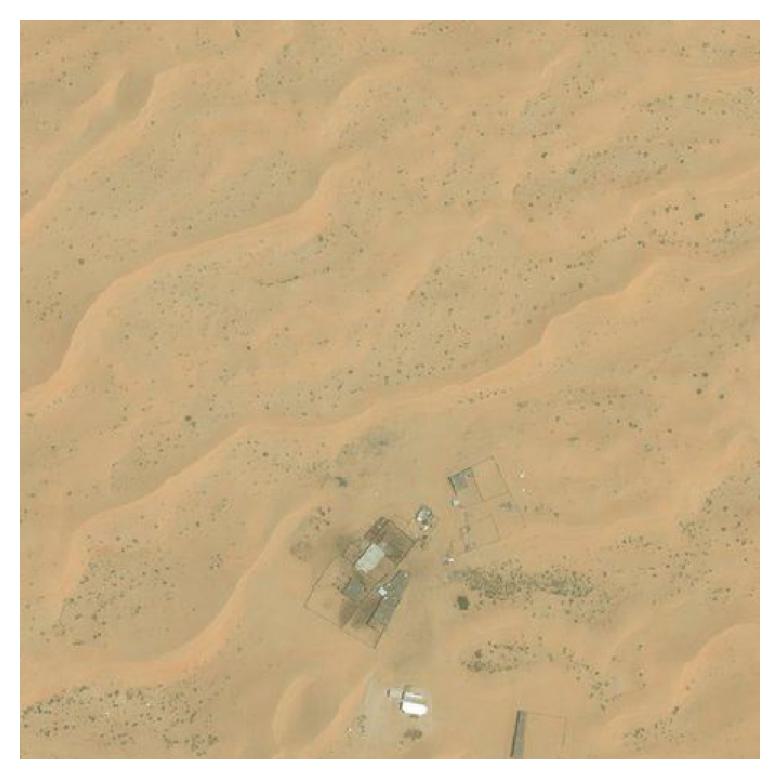}
\end{subfigure}
    \rotatebox{90}{\hspace{2.0em}Thermal}
    \begin{subfigure}[b]{0.15\textwidth}
    \includegraphics[width=\textwidth]{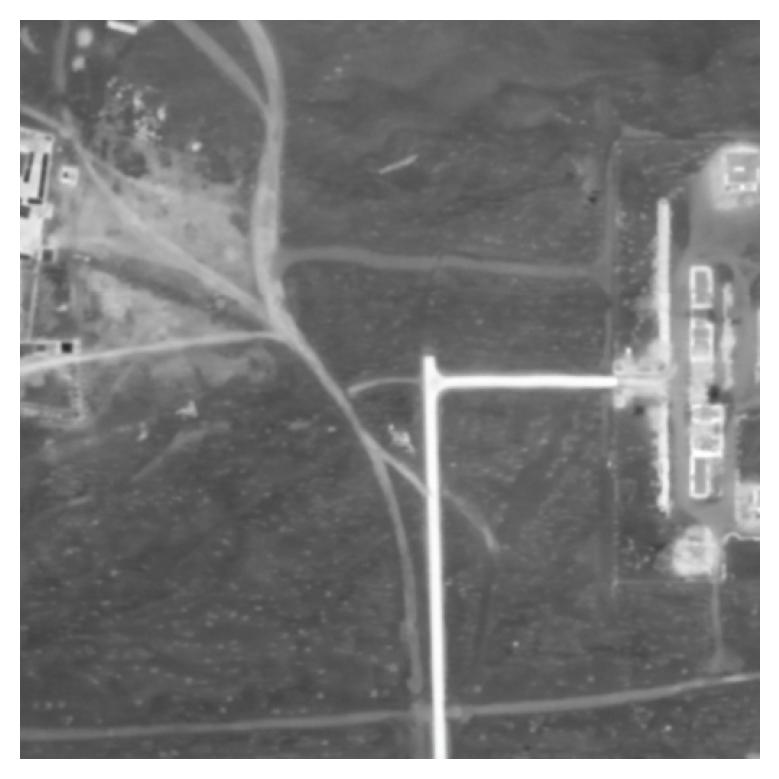}
\end{subfigure}
\begin{subfigure}[b]{0.15\textwidth}
    \includegraphics[width=\textwidth]{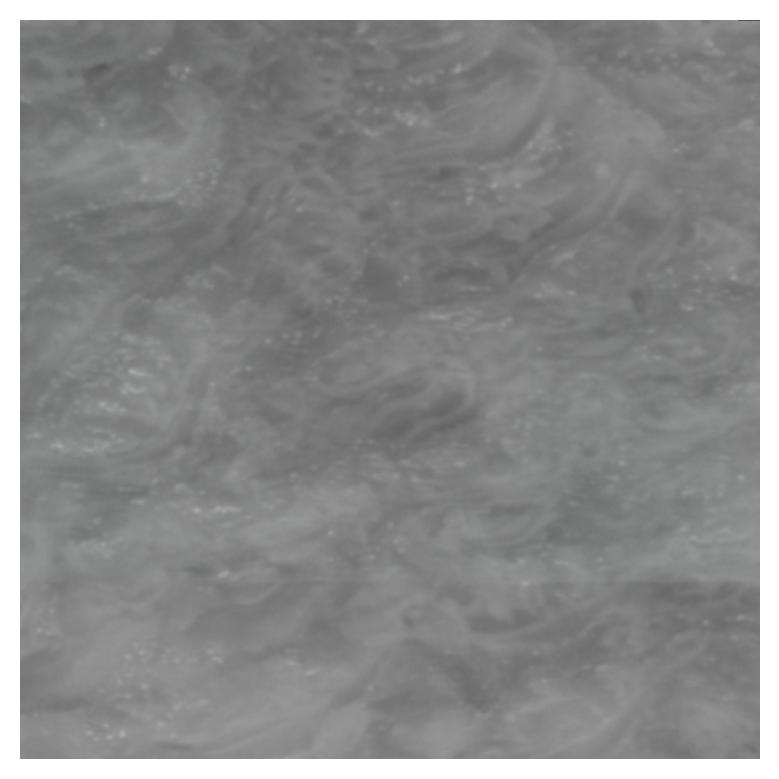}
\end{subfigure}
\begin{subfigure}[b]{0.15\textwidth}
    \includegraphics[width=\textwidth]{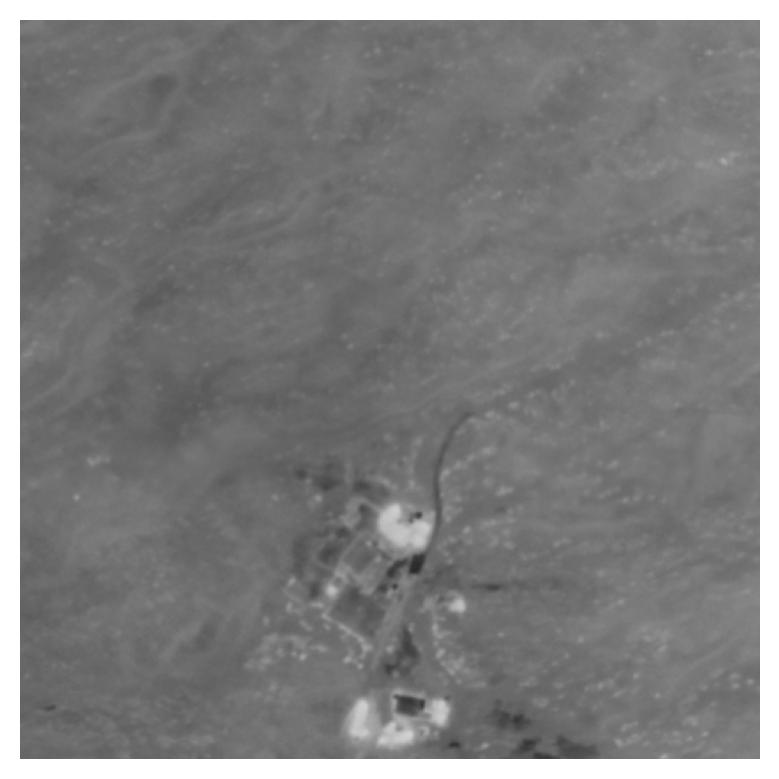}
\end{subfigure}
    \rotatebox{90}{\hspace{1.5em}CE Thermal}
    \begin{subfigure}[b]{0.15\textwidth}
    \includegraphics[width=\textwidth]{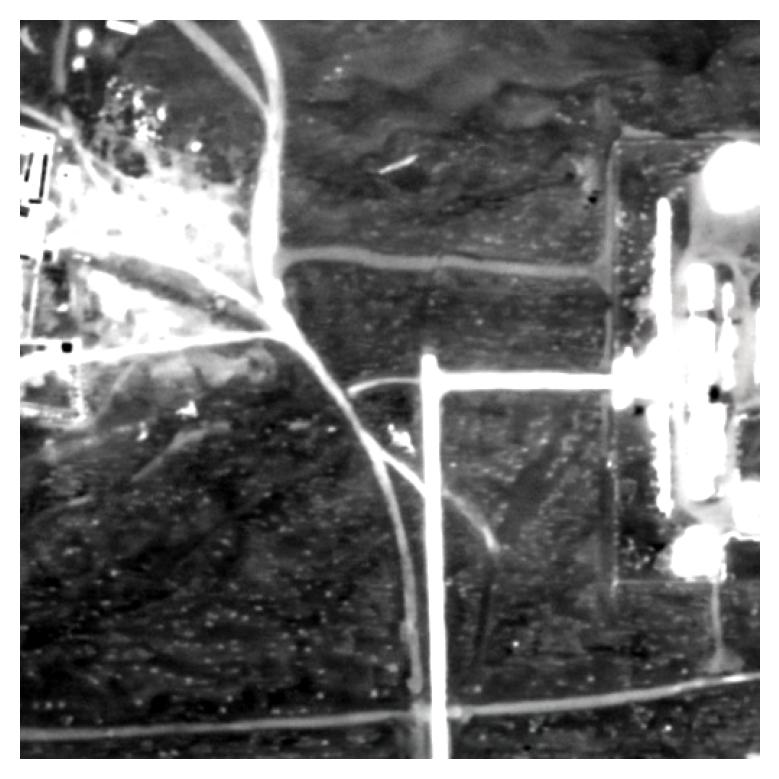}
\end{subfigure}
\begin{subfigure}[b]{0.15\textwidth}
    \includegraphics[width=\textwidth]{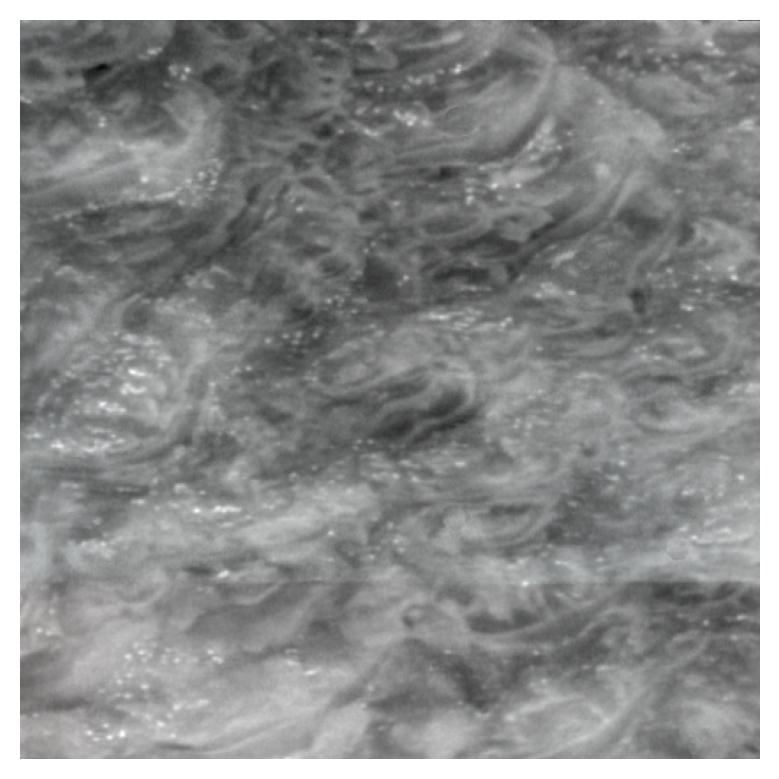}
\end{subfigure}
\begin{subfigure}[b]{0.15\textwidth}
    \includegraphics[width=\textwidth]{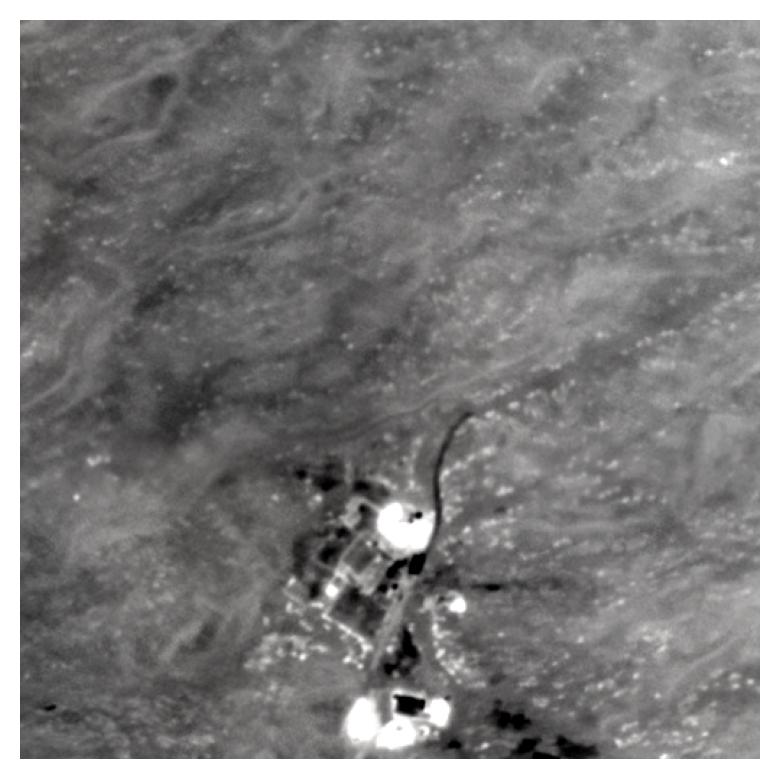}
\end{subfigure}
    \caption{Examples of satellite, raw thermal images, and thermal images with contrast enhancement (CE) in Section~\ref{imp}. The left images depict roads and buildings. The middle column has indistinct self-similar thermal features in deserts. The right column shows the presence of farms.}
    \label{fig:contrast}
\end{figure}
\subsection{Datasets}\label{dataset}
In order to collect thermal aerial data, we used FLIR's Boson thermal imager (8.7 mm focal length, 640p resolution, and $50^\circ$ horizontal field of view)\footnote{\url{https://www.flir.es/products/boson/}}. The collected images are nadir at approx. 1m/px spatial resolution. We performed six flights from 9:00 PM to 4:00 AM and label this dataset as \textbf{Boson-nighttime}, accordingly. To create a single map, we first run a structure-from-motion (SfM) algorithm to reconstruct the thermal map from multiple views. Subsequently, orthorectification is performed by aligning the photometric satellite maps with thermal maps at the same spatial resolution. The ground area covered by Boson-nighttime measures $33~\text{km}{^2}$ in total. The most prevalent map feature is the desert, with small portions of farms, roads, and buildings (Fig. \ref{fig:contrast}). The low-contrast self-similar thermal features in the desert regions have made it difficult for geo-localization at a finer scale (Fig. \ref{fig:contrast}).


The Bing satellite map\footnote{Bing satellite imagery is sourced from Maxar: \url{https://www.bing.com/maps/aerial}} is cropped in the corresponding area as our satellite reference map. We tile the thermal map into $512\times512$~px thermal image crops with a stride of $35$ px. Each thermal image crop pairs with the corresponding satellite image crop. Areas covered by three flights of Boson-nighttime are used for training and validation. The remaining areas, covered by the other three flights are used for testing. The train/validation/test splits for Boson-nighttime are $10256$/$13011$/$26568$ pairs of satellite and thermal image crops, respectively. 

As a way to train TGM, we remove the images with invalid regions on the SfM-reconstructed thermal map. The gaps between captured thermal images cause these invalid regions. We find that TGM training with these images can create artifacts on the generated thermal images. SGM training still employs images with invalid regions as a way of adding noise to the data. The generated dataset has $79950$ pairs of satellite and generated thermal images.

\begin{table*}[ht]
    \centering
    \smallskip
    \caption{The results of different settings of CE, DANN loss, and Generated Dataset. The \textbf{bold} value shows the best result and the \underline{underline} value shows the second-best result. The arrows with metrics show the direction of good values.}
    \begin{tabular}{cccccccccc}
    \toprule
         Backbone & CE & DANN & DANN Only Positive & Generated Dataset & $R@1\uparrow$ & $R@5\uparrow$ & $R_{512}@1\uparrow$ & $R_{512}@5\uparrow$ & $L^{512}_{2}(m)\downarrow$\\\midrule
         \multirow{6}{5em}{ResNet-18} & & & & & \underline{69.3} & \underline{81.8} & 81.1 & 92.0 & \underline{58.9}\\
          & & \checkmark & & & 61.1 & 74.5 & 75.5  & 87.7 & 77.8\\
          & & \checkmark & \checkmark & & 67.4 & 80.2 & 81.4 & 92.1 & 59.8\\
          & \checkmark & & & & 68.0 & 80.5  & \underline{82.1} & \underline{92.2}& \underline{58.9}\\
          & \checkmark & \checkmark & & & 60.1 & 74.5 & 76.1 & 88.9 & 75.8\\
          & \checkmark & \checkmark & \checkmark & & \textbf{72.1} & \textbf{83.9} & \textbf{84.5} & \textbf{93.7} & \textbf{52.2}\\\midrule
         \multirow{6}{5em}{ResNet-18} & & & & \checkmark & 84.8 & 90.8 & \underline{95.5} & \underline{98.7} & \underline{19.0}\\
         & & \checkmark & & \checkmark & 68.9 & 77.0 &82.3 & 89.5 & 65.9\\
         & & \checkmark & \checkmark &  \checkmark & 82.0 & 88.8 & 93.7 & 97.9 & 24.8
\\
         & \checkmark & & &\checkmark & \underline{87.0} &  \underline{91.8} & 94.2 & 97.6 & 24.2\\
         & \checkmark & \checkmark & & \checkmark & 81.2 & 88.7 & 90.3 & 95.7 & 34.6\\
         & \checkmark & \checkmark & \checkmark & \checkmark & \textbf{92.1}& \textbf{96.9}& \textbf{96.5}& \textbf{99.1} & \textbf{14.7}\\
         \bottomrule
    \end{tabular}
    \label{tab:main}
\end{table*}

\subsection{Implementation Details}\label{imp}
For the training of TGM, the size of the input image crops is $512\times512$. To improve generative quality, we upsample satellite and thermal input images to $1024\times1024$. The generated thermal images are then downsampled to $512\times512$ for SGM training efficiency. The number of training epochs is $40$, and the batch size is $8$. Following the pix2pix settings \cite{pix2pix}, we use U-Net \cite{unet} architecture for generator $G$ and Adam optimizer \cite{adam} for training. The learning rate is set to $0.0002$, and a linear learning rate decay scheduler is applied after $20$ epochs. We remove dropout layers in the generator $G$ of TGM for deterministic thermal results.

In the SGM, each batch consists of a query, a positive, and ten negative samples following the deep-VG-benchmark settings~\cite{Berton_CVPR_2022_benchmark}. Ten triplets are built by the query, positive samples, and one of the negative samples in one batch. We use a batch size of $4$. Adam optimizer \cite{adam} is used and the learning rate is set to $1\times10^{-4}$.  We mainly use ResNet-18 \cite{resnet} as the backbone (feature extractor $F$ in Fig.~\ref{fig:framework}). We use \textit{partial mining} \cite{Berton_CVPR_2022_benchmark} for triplet mining. We run for $100$ epochs, and each epoch iterates $5000$ queries. Each epoch caches $5000$ database embeddings and runs triplet mining to find the positives and negatives. The positives are within a $35$-meter radius of the queries, and the negatives are selected among those further than $50$ meters. The output dimension of descriptors is controlled to $4096$. The pipelines are implemented using PyTorch \cite{NEURIPS2019_9015}. The embedding extraction time is $12.84$~ms with inference batch size $=1$ and on an NVIDIA RTX2080Ti GPU. The matching time is $1.20$~ms with a database of $26568$ satellite embeddings.

To enhance the indistinct thermal features, we have the option to do contrast enhancement (CE) on thermal images. We use the linear scaling contrast adjustment algorithm with contrast factor $=3$. The parameter is empirically tuned by investigating the histogram of the thermal dataset. The enhanced examples are shown in Fig. \ref{fig:contrast}.

\subsection{Metrics} \label{metrics}
To evaluate the thermal geo-localization performance, we use the following metrics
\begin{itemize}
    \item Recall@$N$ ($R@N$): It measures the percentage of the query images from which the top-$N$ retrieved database images are within $50$ meters.
    \item Recall@$N$ with prior location threshold $d$ ($R_d@N$): It is $R@N$ with the search region limited to a radius of $d$ meters from the queries. This metric assumes that we coarsely know a potential search region, given the current flight's last location and the UAV's motion estimation. We show the results of $R_{512}@1$ and $R_{512}@5$.
    \item $L_{2}$ distance error with prior location threshold $d$ ($L^d_{2}$): It measures the $L_2$ distance (meter) from the queries to the estimated position from top-$N$ retrieved database images within a radius of $d$ meters from the queries. We use a simple position estimation algorithm that uses the Top-1 retrieved position, while more stable results can be obtained if more retrieved candidates are considered. We show the results of $L^{512}_{2}$.
\end{itemize}
\section{Results}~\label{sec:resutls}
The results of our proposed framework are shown in Tables \ref{tab:main}-\ref{tab:TGM}. We investigate the effect of contrast enhancement (CE) (Section~\ref{sec:contrast}), DANN settings (Section~\ref{DANN_set}), and the impact of the generated dataset from TGM (Section~\ref{gen}) in Table~\ref{tab:main} and \ref{tab:TGM}. We also show the visualized TGM generative results and SGM geo-localization results (Fig. \ref{fig:generative} and Fig. \ref{fig:geo}) in Generated Dataset Analysis (Section~\ref{gen}) and Visualized Geo-localization Results (Section~\ref{sec:geo}).

\subsection{Contrast Enhancement Analysis}\label{sec:contrast}

The raw thermal images are low-contrast with indistinct self-similar features, which may impact the geo-localization performance. We show the results in the upper part of Table~\ref{tab:main} and show as well the impact of Contrast Enhancement (CE) on thermal images. We compare the models without generated dataset and find that the models with CE have higher $R_{512}@1$ and $R_{512}@5$ and lower or equal $L^{512}_{2}$ than those without CE. This reveals that using enhanced thermal images can boost geo-localization performances for low-contrast thermal features.

\textbf{Discussion.} The models with CE sometimes show worse recall and distance error metrics than those without CE. This also happens when the models use generated datasets (lower part of Table \ref{tab:main}). The exception is models with CE and DANN (only positive). From Fig. \ref{fig:contrast}, we observe that the enhancement distorts thermal features at the building region. These facts imply that the enhanced thermal image may mismatch a wrong satellite image for distorted thermal features and result in localization errors.

\subsection{DANN Analysis}\label{DANN_set}

In the upper part of Table~\ref{tab:main}, we show the results with or without DANN. \textit{DANN only positive} means no negative samples are considered in the DANN loss. We look into the effectiveness of DANN loss and the necessity to remove negatives from DANN loss. We find that DANN loss with negatives always lowers the recall performance and increases $L_2$ distance error, which practically supports our assumption that DANN loss can conflict with Triplet margin loss in Section~\ref{sec:SGM}. We also discover that DANN loss without negative samples typically works with enhanced thermal images. The model with CE and DANN (only positive) shows the best geo-localization performance among the models.

\begin{table}[]
    \centering
    \caption{The results of using generated dataset by different settings of CE and $\lambda_1$ in training TGM.}
    \begin{tabular}{ccrrrr}
    \toprule
         Backbone & CE & $\lambda_1$ & $R@1\uparrow$ & $R_{512}@1\uparrow$ & $L^{512}_{2}(m)\downarrow$\\\midrule
         \multirow{4}{5em}{ResNet-18} & & 10.0 & 7.2 & 20.4 & 299.6\\
          & \checkmark & 10.0 & \underline{90.5} & \underline{95.9}& \underline{17.1}\\
          & & 100.0 & 82.0& 93.7
& 24.8\\
          & \checkmark & 100.0 & \textbf{92.1} & \textbf{96.5} & \textbf{14.7}\\
         \bottomrule
    \end{tabular}
    \label{tab:TGM}
\end{table}

\textbf{Discussion.} Our results reveal that the DANN method can sometimes worsen certain metrics. We observe this trend across models that leverage a generated dataset and those that do not. These results suggest that incorporating both CE and DANN (only positive) is crucial in our task.

\subsection{Generated Dataset Analysis}\label{gen}
To exploit the considerable number of unpaired satellite images, we use TGM to synthesize thermal images and build a generated paired dataset. In the lower part of Table~\ref{tab:main}, we observe that models with the generated dataset notably improve the recall performance and $L_2$ error. Our Best Model (ResNet-18 with CE, DANN only positive, and the generated dataset) exhibits accurate localization results with $R@1>90$ and $R@5>95$ and $L^{512}_{2}<15$, which significantly outperforms Baseline Model (ResNet-18). These findings highlight the ability of synthesized datasets to improve model performance when paired data is limited. Overall, these results provide compelling evidence for the effectiveness of our approach to thermal geo-localization. Moreover, we analyze the training settings of TGM, which primarily influence the quality of the generated dataset. In Table~\ref{tab:TGM}, we investigate the influence of utilizing CE and adjusting $\lambda_1$. The default setting is with DANN (only positive). The results show that CE and $\lambda_1$ are the first and second most important factors determining the thermal geo-localization performance.

\begin{figure}
    \centering
    \includegraphics[width=0.49\textwidth]{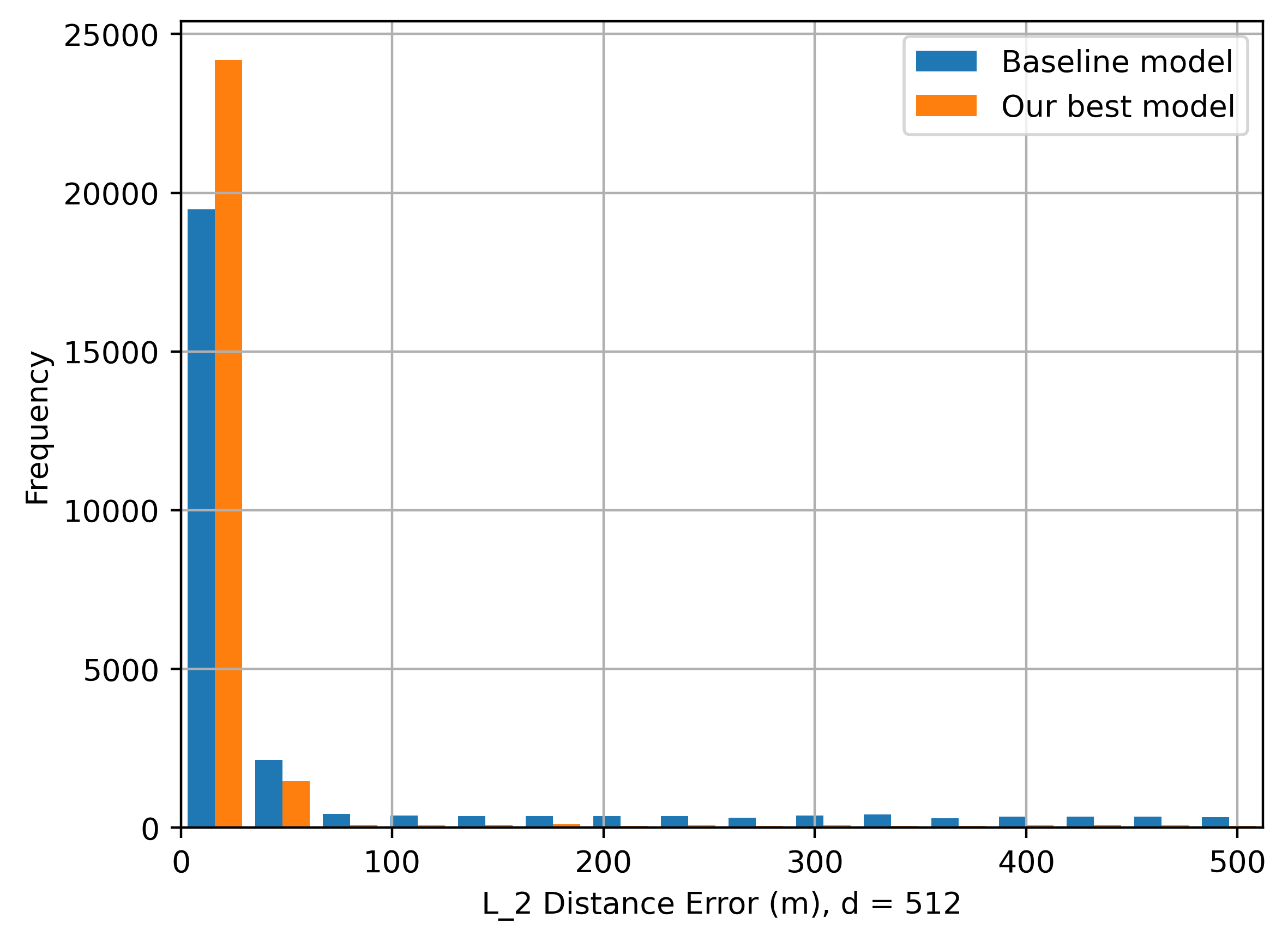}
    \caption{The histogram of $L^{512}_2$ Distance Error of Baseline Model (ResNet-18) and Our Best Model (ResNet-18 with CE, DANN only positive, and generated dataset).}
    \label{fig:hist}
\end{figure}

In Fig. \ref{fig:hist}, we show the histogram of $L_2$ distance error of the Baseline Model and Our Best Model. The histogram shows that Our Best Model outperforms Baseline Model by showing more low-error cases. Additionally, Baseline Model exhibits considerable instances of localization failure with $L_2$ error larger than 100 meters, while Our Best Model shows fewer such cases.

\begin{figure}[]
\smallskip
\smallskip
    \centering
\rotatebox{90}{\scriptsize\hspace{2.7em}Satellite}
\begin{subfigure}[b]{0.11\textwidth}
    \includegraphics[width=\textwidth]{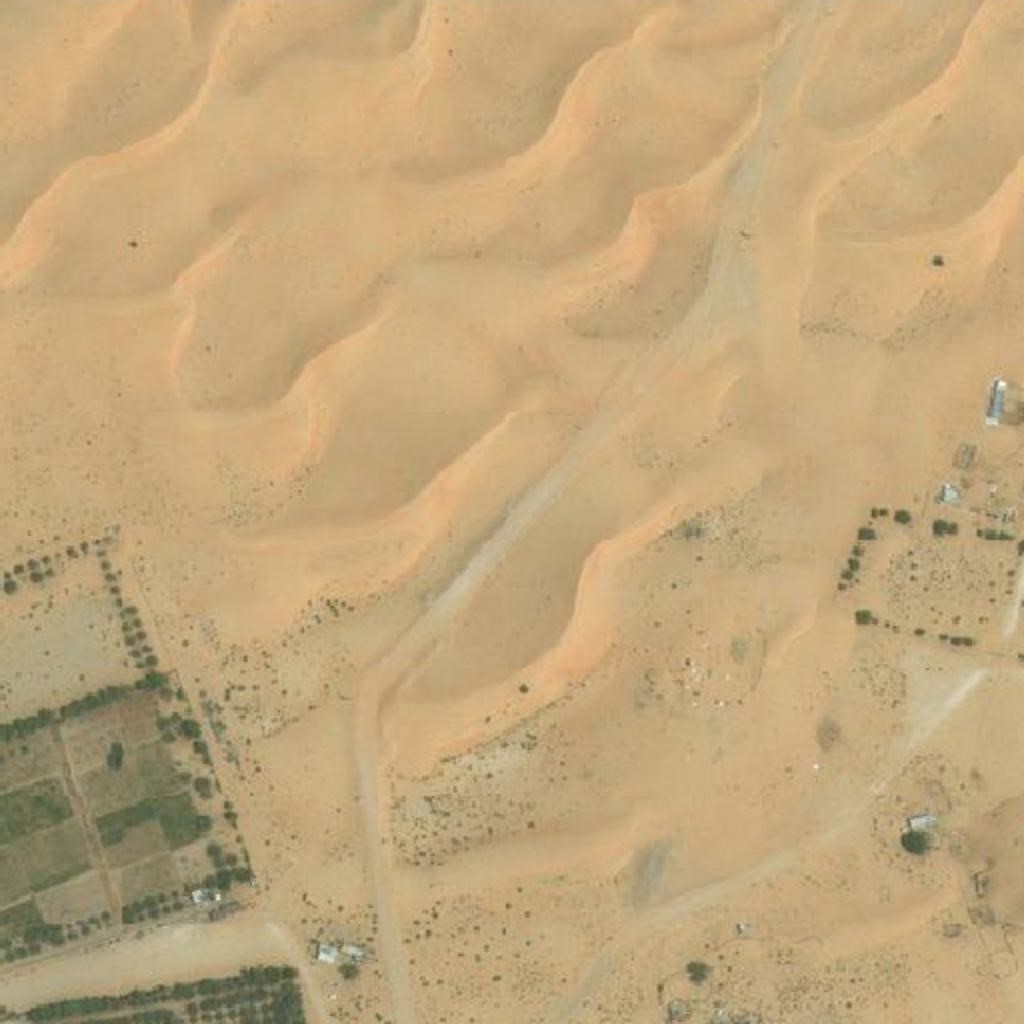}
    \vspace{-0.8\baselineskip}
\end{subfigure}
\begin{subfigure}[b]{0.11\textwidth}
    \includegraphics[width=\textwidth]{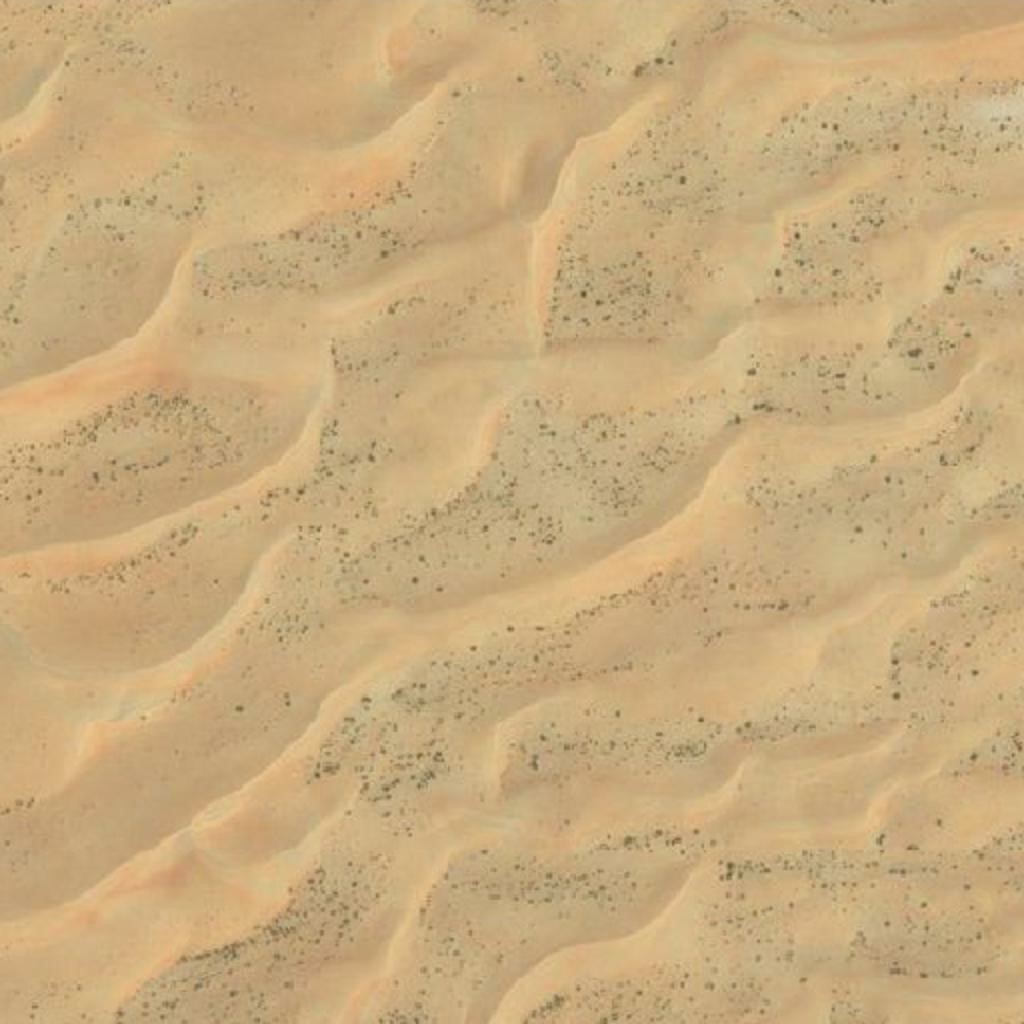}
    \vspace{-0.8\baselineskip}
\end{subfigure}
\begin{subfigure}[b]{0.11\textwidth}
    \includegraphics[width=\textwidth]{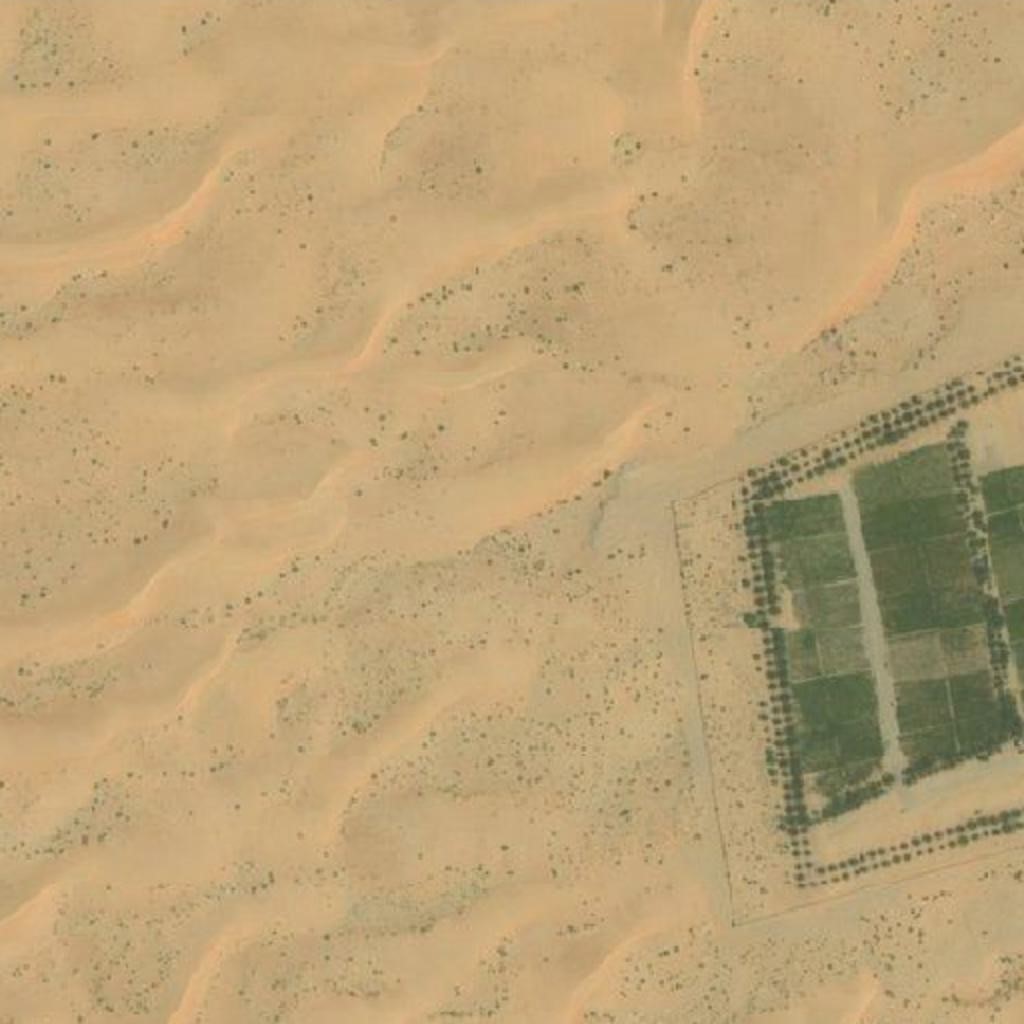}
    \vspace{-0.8\baselineskip}
\end{subfigure}
\begin{subfigure}[b]{0.11\textwidth}
    \includegraphics[width=\textwidth]{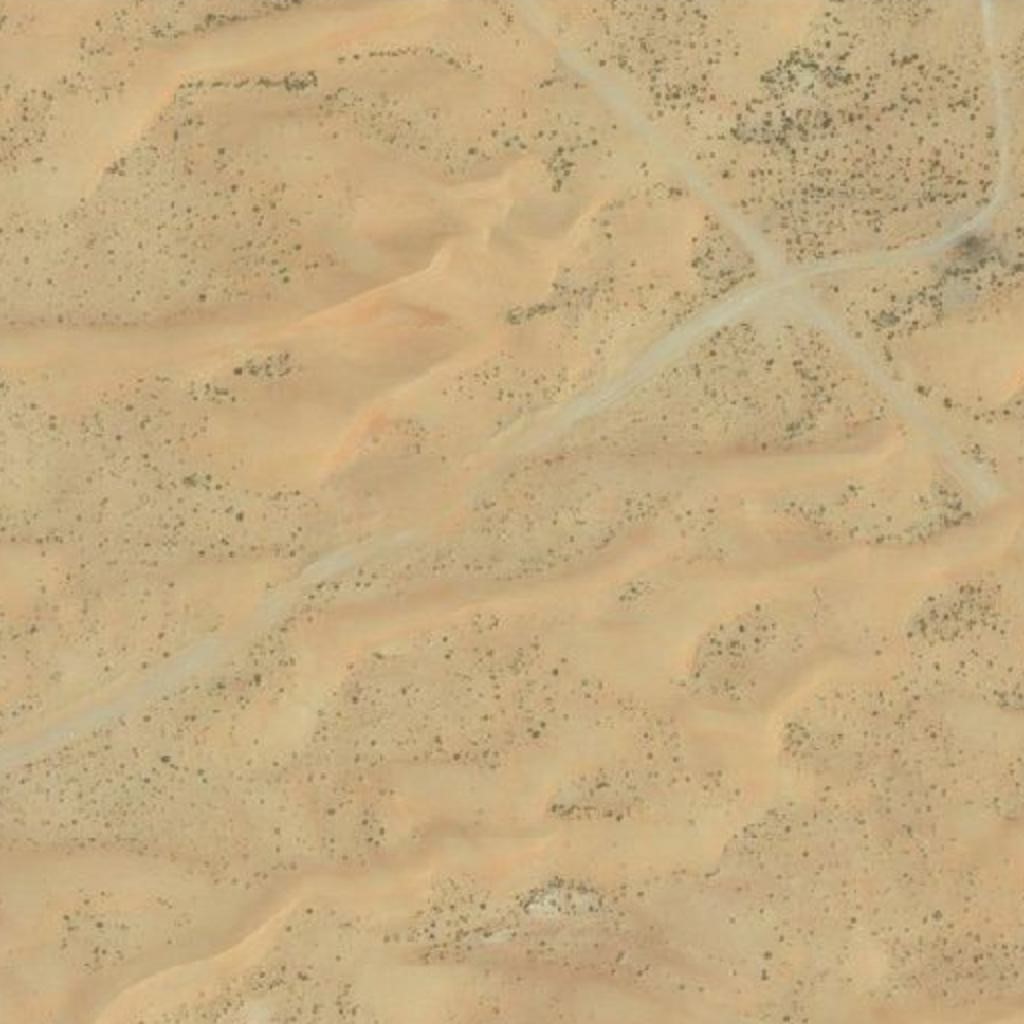}  
    \vspace{-0.8\baselineskip}
\end{subfigure}
\rotatebox{90}{\scriptsize\hspace{2em}GT Thermal}
\begin{subfigure}[b]{0.11\textwidth}
    \includegraphics[width=\textwidth]{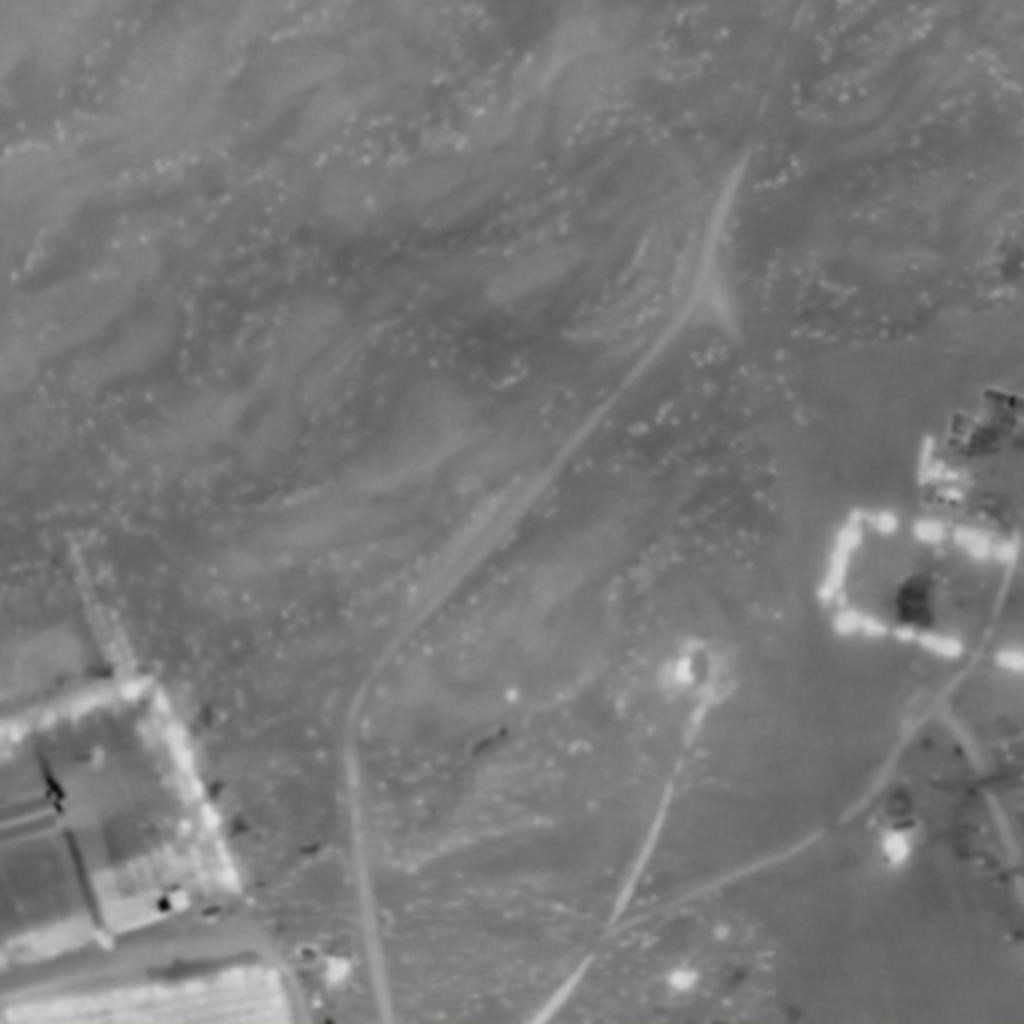}
    \vspace{-0.8\baselineskip}
\end{subfigure}
\begin{subfigure}[b]{0.11\textwidth}
    \includegraphics[width=\textwidth]{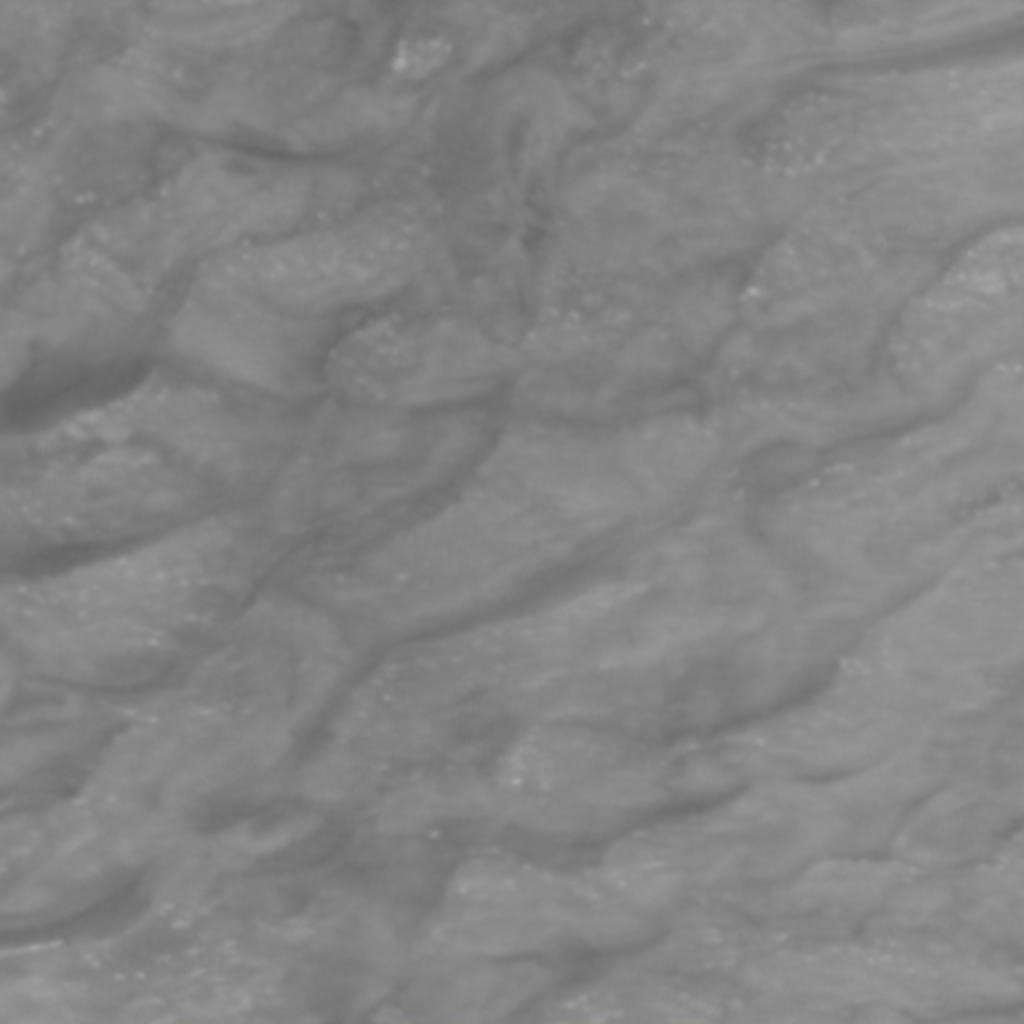}
    \vspace{-0.8\baselineskip}
\end{subfigure}
\begin{subfigure}[b]{0.11\textwidth}
    \includegraphics[width=\textwidth]{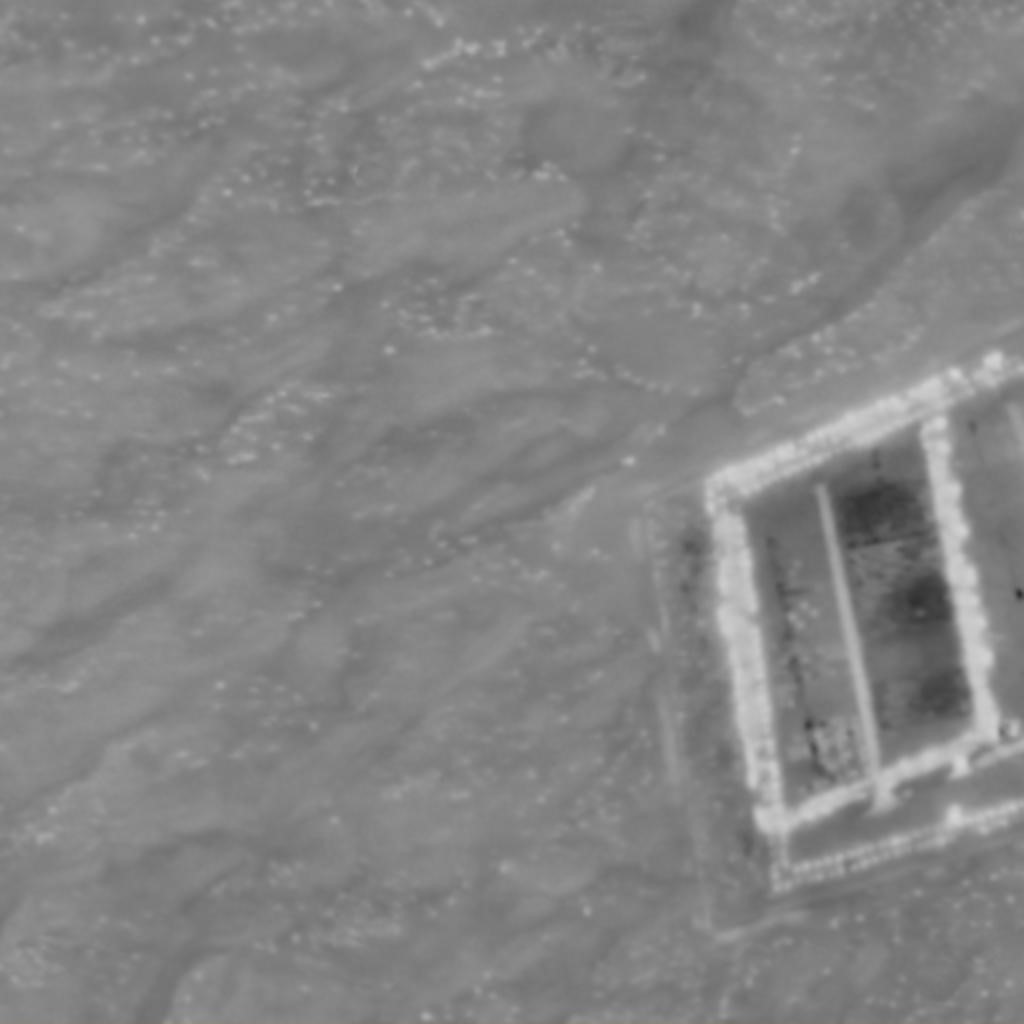}
    \vspace{-0.8\baselineskip}
\end{subfigure}
\begin{subfigure}[b]{0.11\textwidth}
    \includegraphics[width=\textwidth]{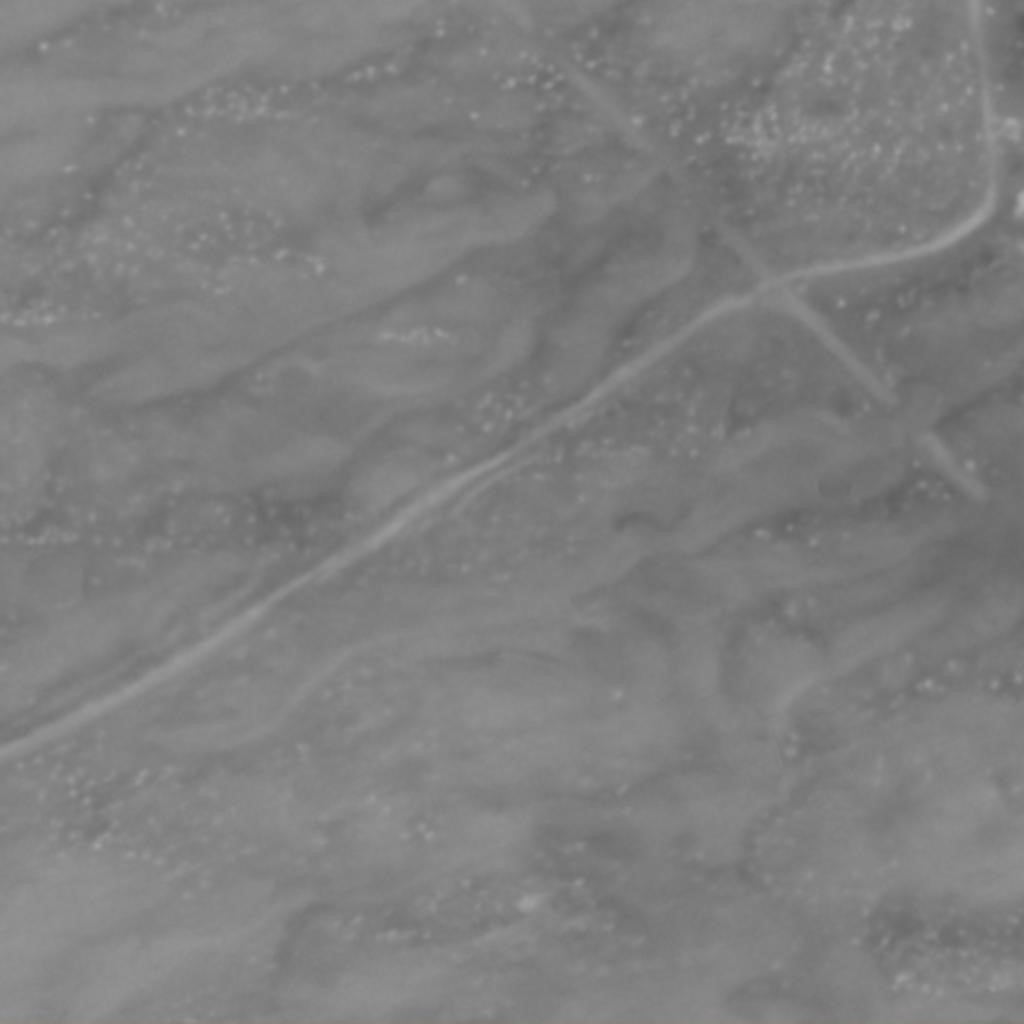}
    \vspace{-0.8\baselineskip}
\end{subfigure}
\rotatebox{90}{\scriptsize\enspace\hspace{1em}Gen. Thermal}
\begin{subfigure}[b]{0.11\textwidth}
    \includegraphics[width=\textwidth]{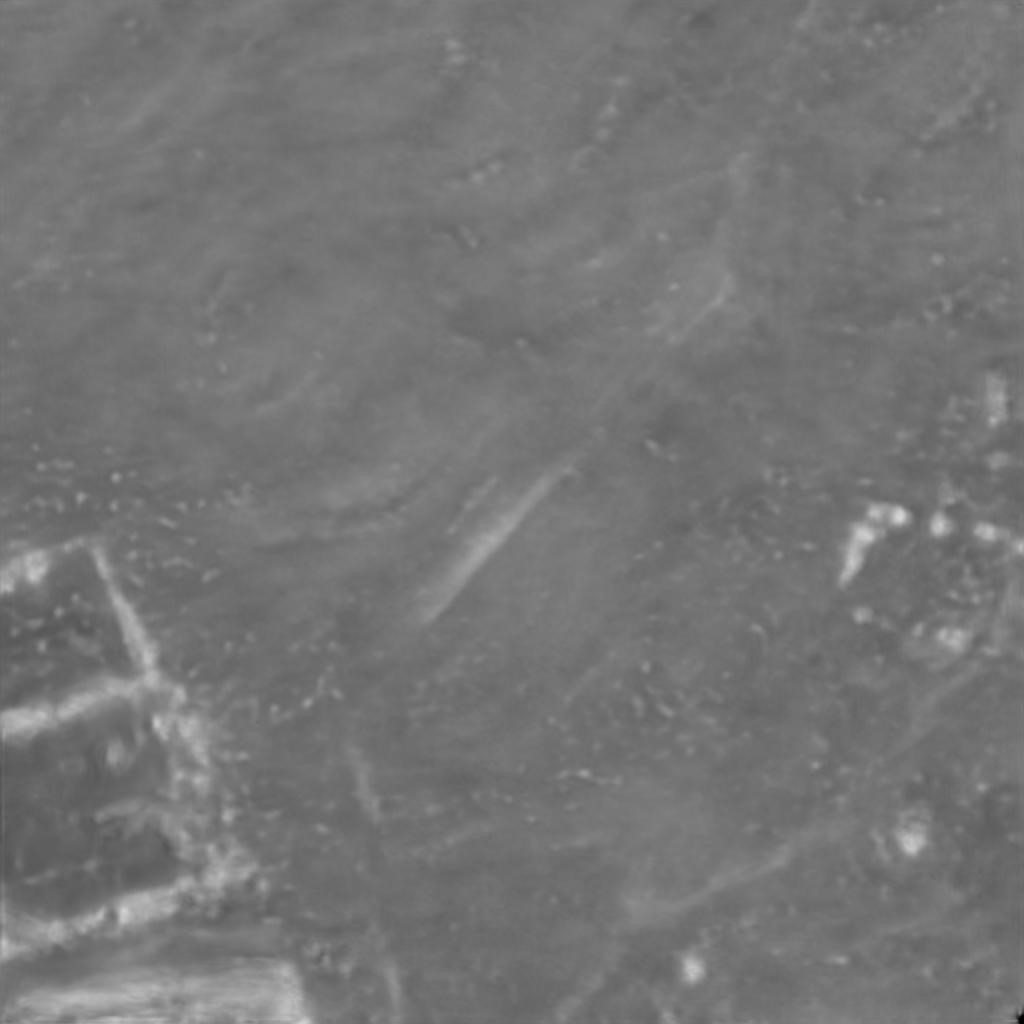}
    \vspace{-0.8\baselineskip}
\end{subfigure}
\begin{subfigure}[b]{0.11\textwidth}
    \includegraphics[width=\textwidth]{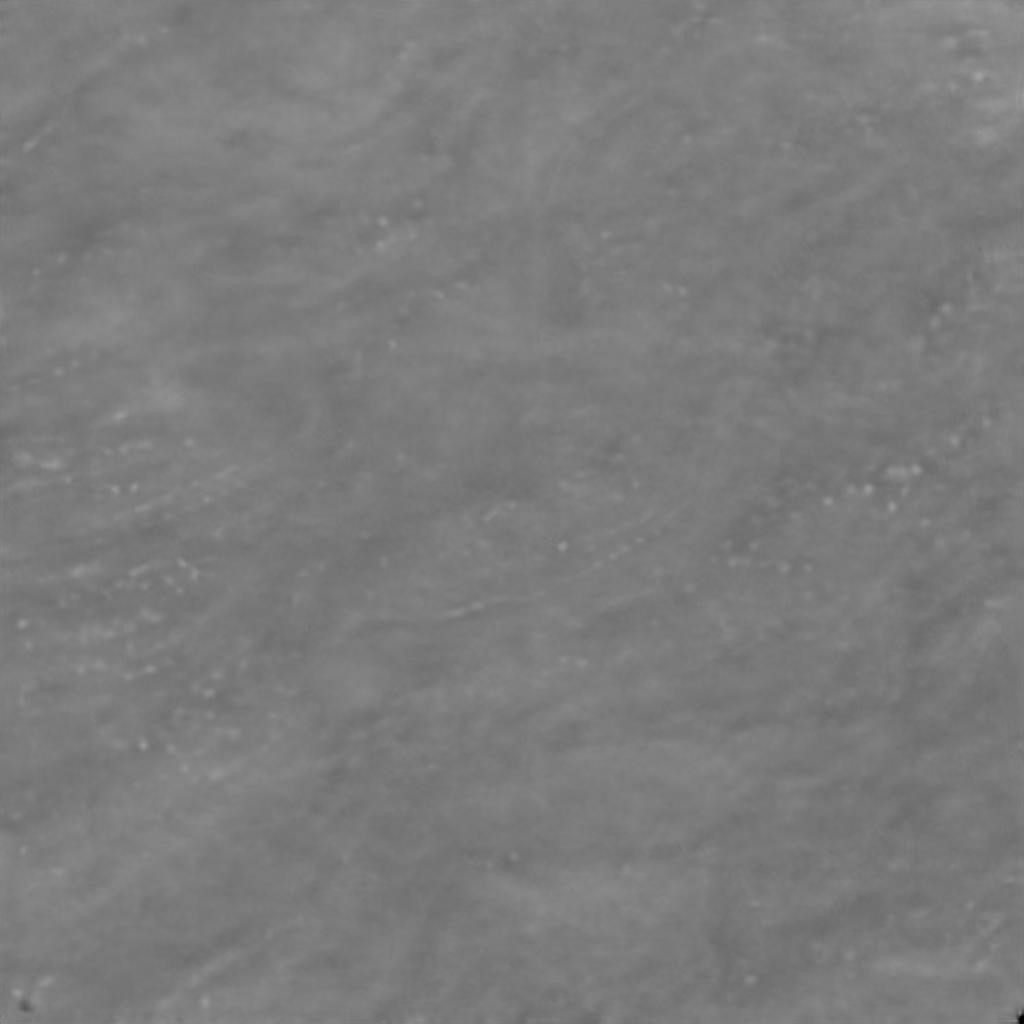}
    \vspace{-0.8\baselineskip}
\end{subfigure}
\begin{subfigure}[b]{0.11\textwidth}
    \includegraphics[width=\textwidth]{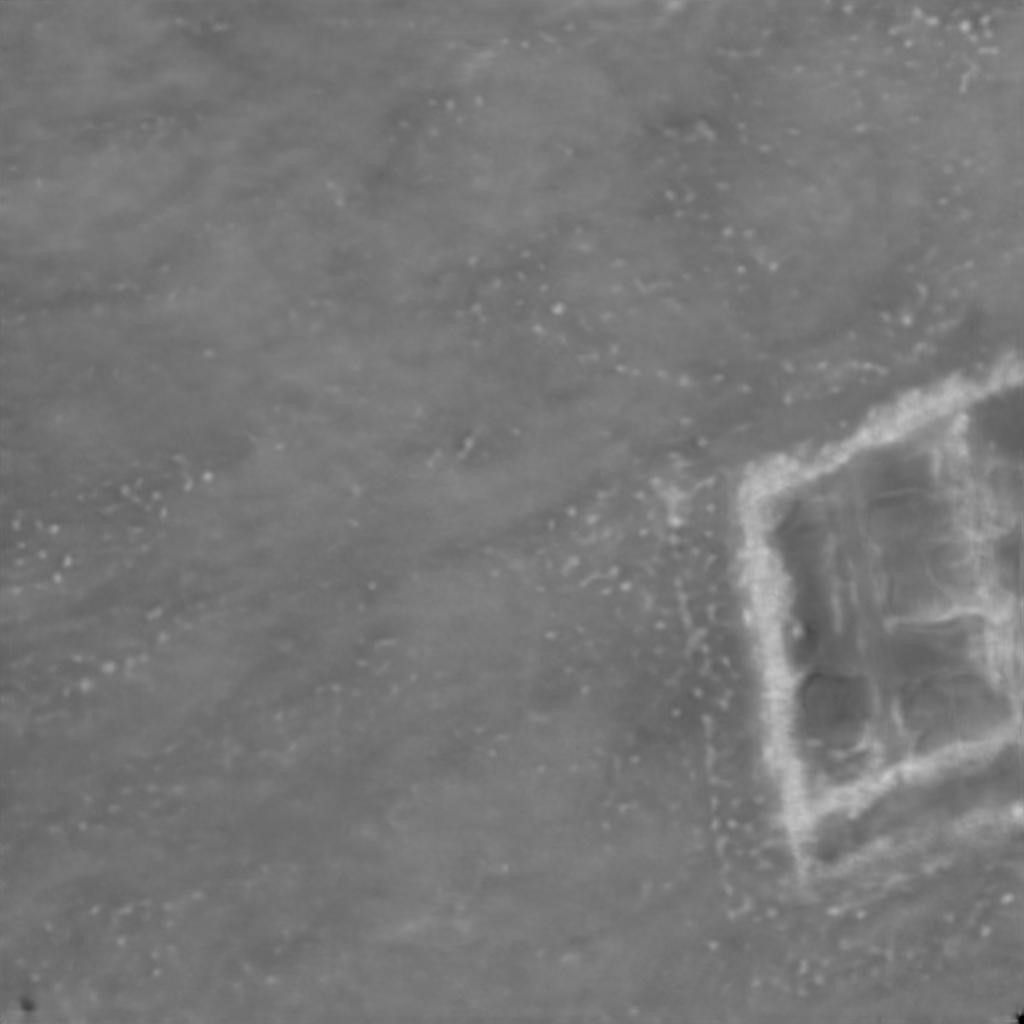}
    \vspace{-0.8\baselineskip}
\end{subfigure}
\begin{subfigure}[b]{0.11\textwidth}
    \includegraphics[width=\textwidth]{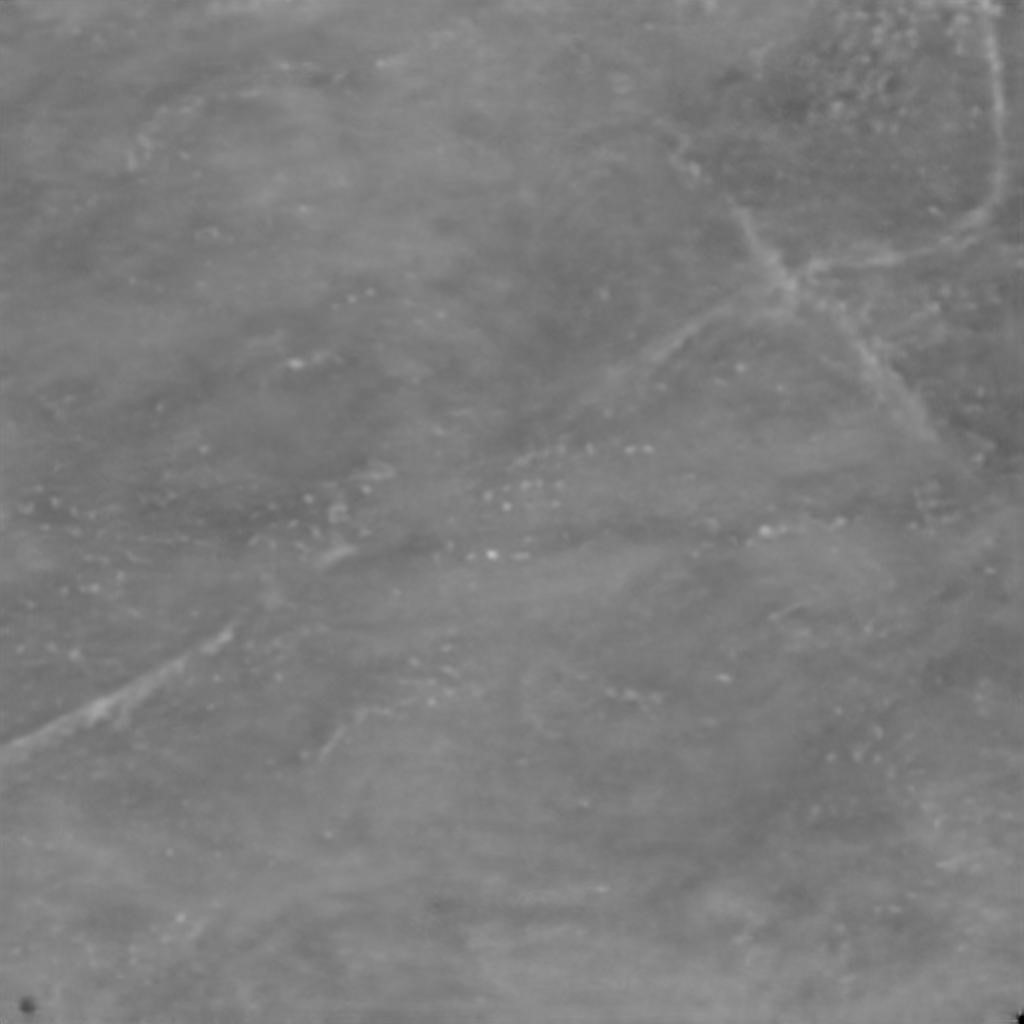}
    \vspace{-0.8\baselineskip}
\end{subfigure}
\rotatebox{90}{\scriptsize\hspace{1em}GT CE Thermal}
\begin{subfigure}[b]{0.11\textwidth}
    \includegraphics[width=\textwidth]{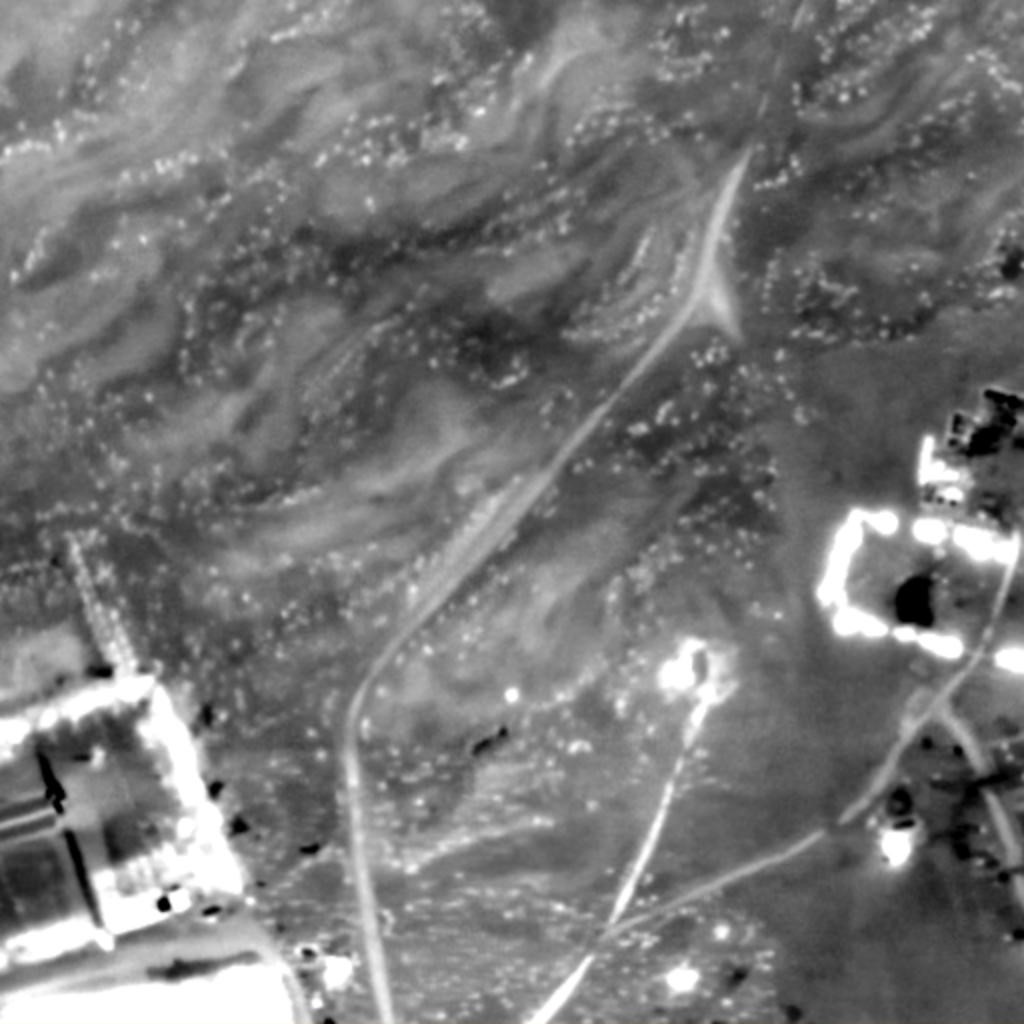}
    \vspace{-0.8\baselineskip}
\end{subfigure}
\begin{subfigure}[b]{0.11\textwidth}
    \includegraphics[width=\textwidth]{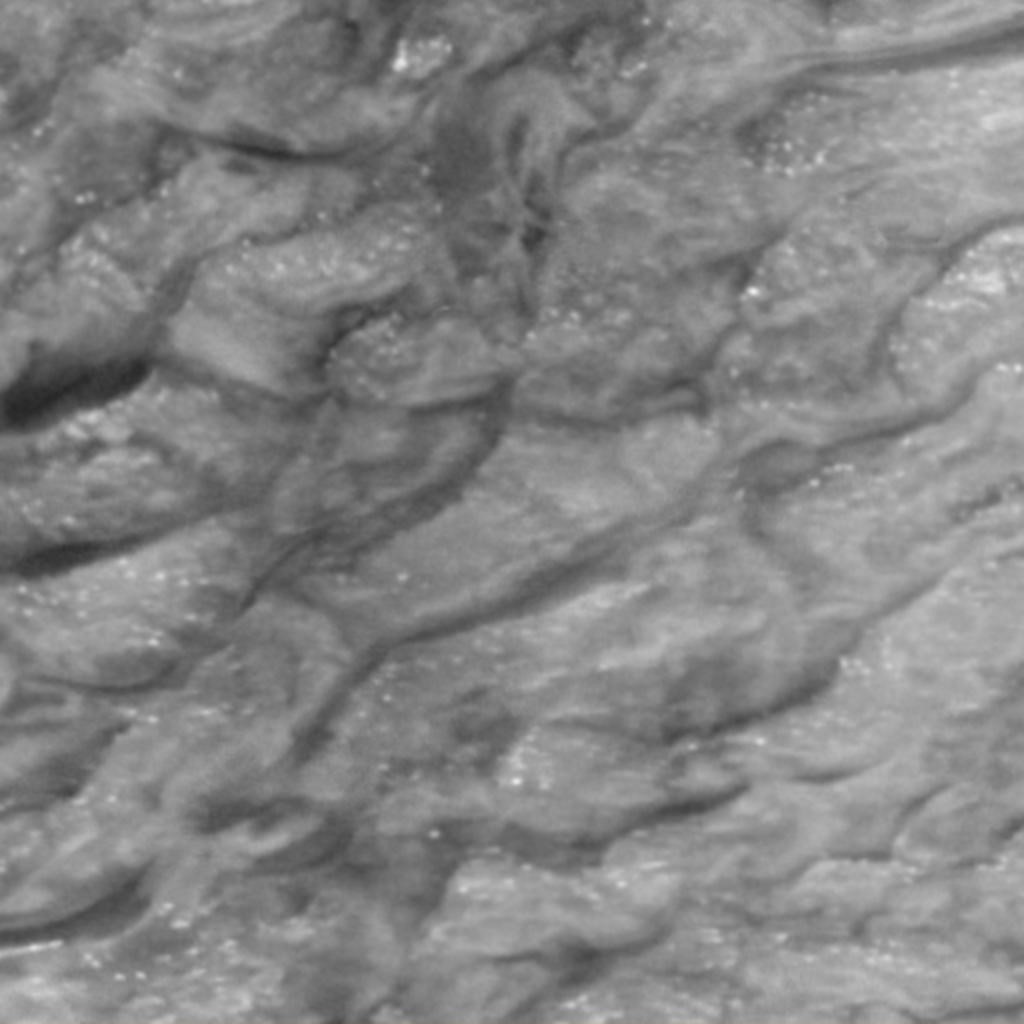}
    \vspace{-0.8\baselineskip}
\end{subfigure}
\begin{subfigure}[b]{0.11\textwidth}
    \includegraphics[width=\textwidth]{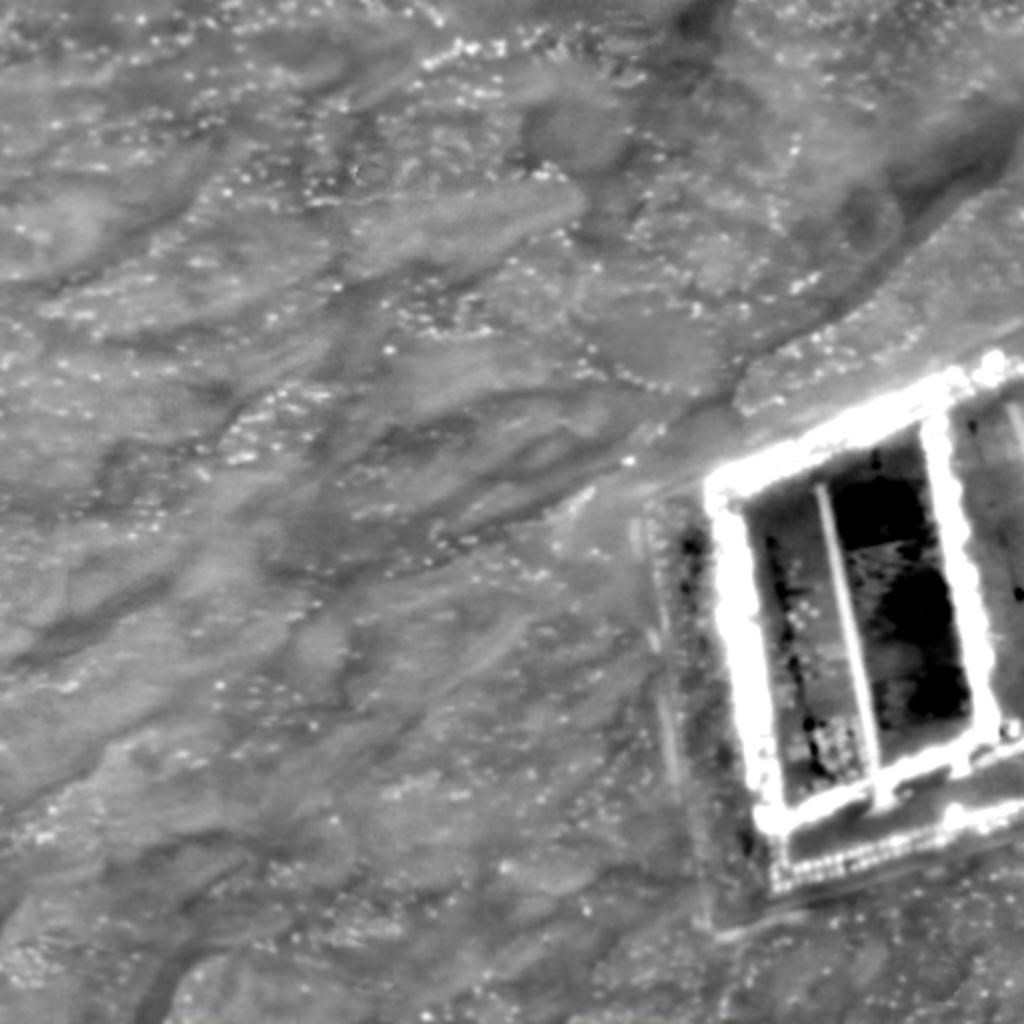}
    \vspace{-0.8\baselineskip}
\end{subfigure}
\begin{subfigure}[b]{0.11\textwidth}
    \includegraphics[width=\textwidth]{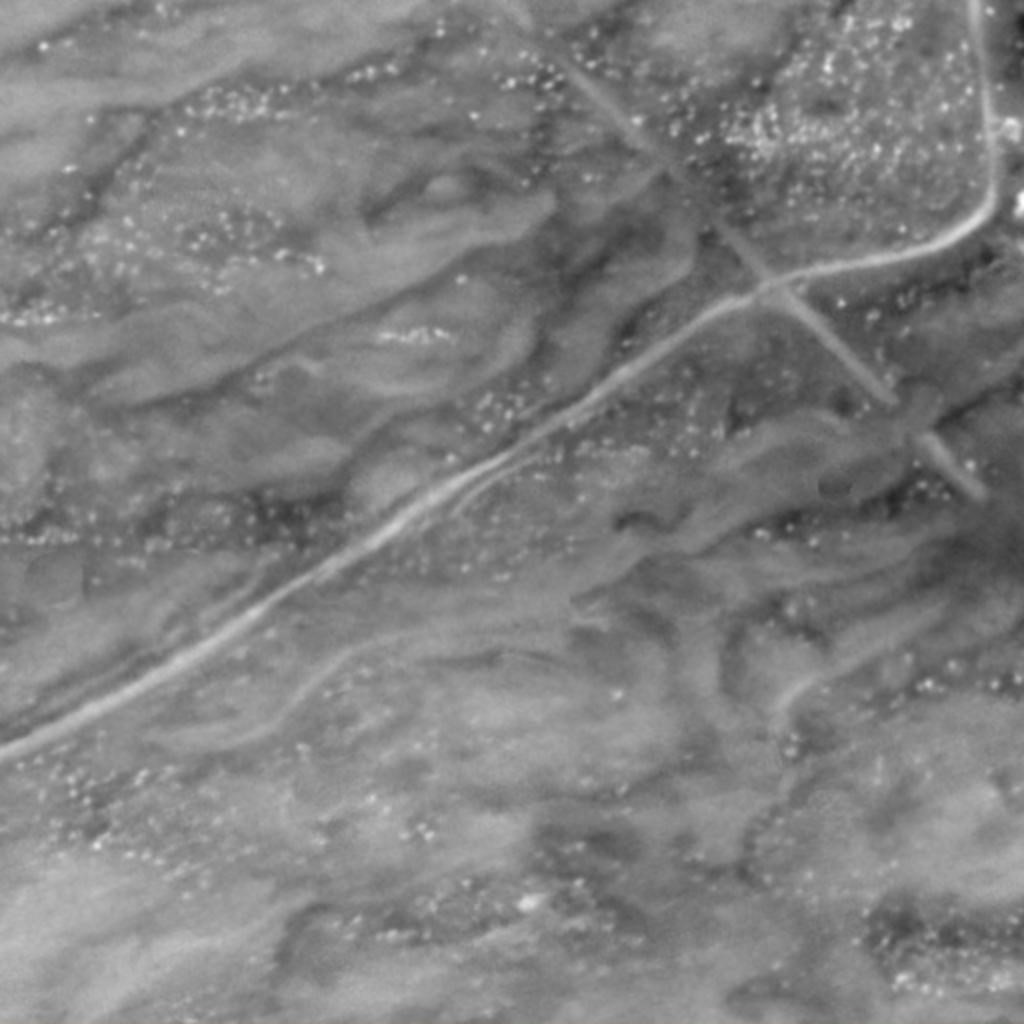}
    \vspace{-0.8\baselineskip}
\end{subfigure}
\rotatebox{90}{\scriptsize \enspace Gen. CE Thermal}
\begin{subfigure}[b]{0.11\textwidth}
    \includegraphics[width=\textwidth]{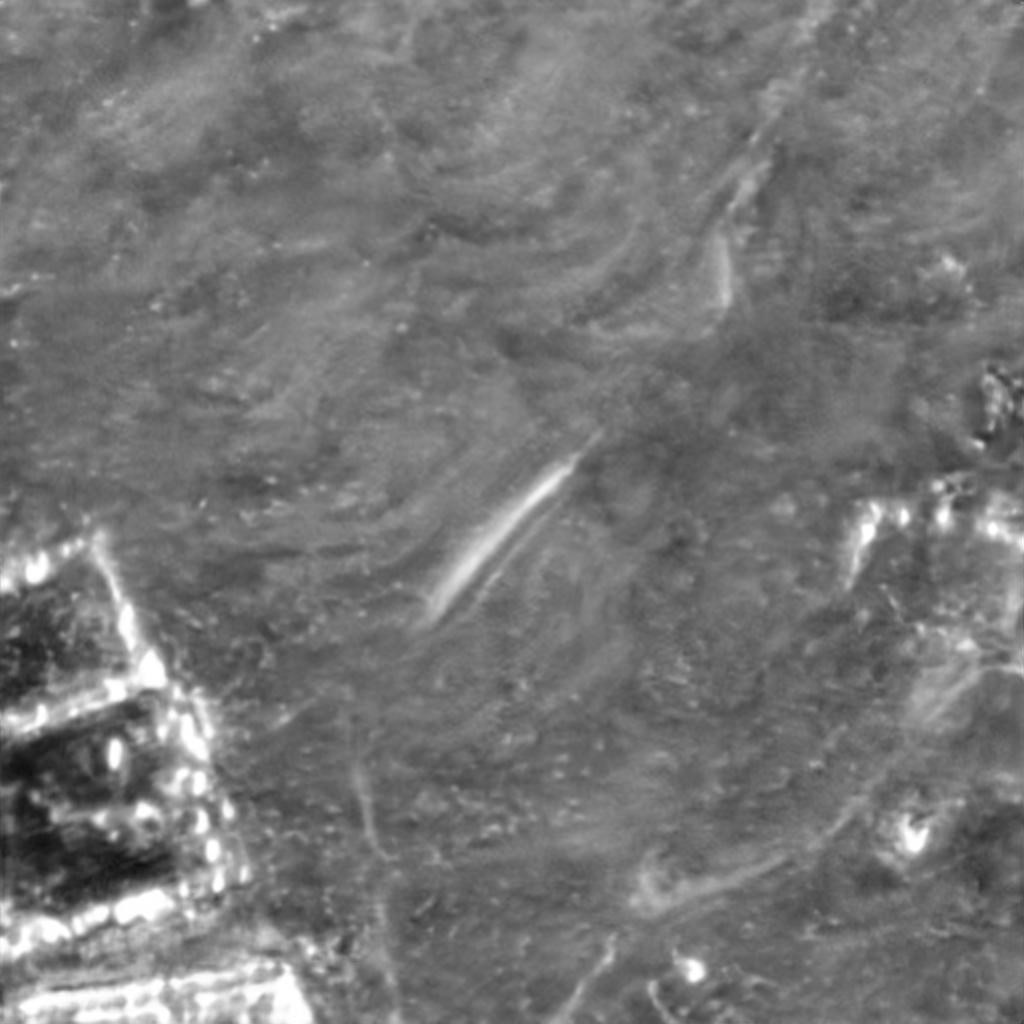}
    \vspace{-0.8\baselineskip}
\end{subfigure}
\begin{subfigure}[b]{0.11\textwidth}
    \includegraphics[width=\textwidth]{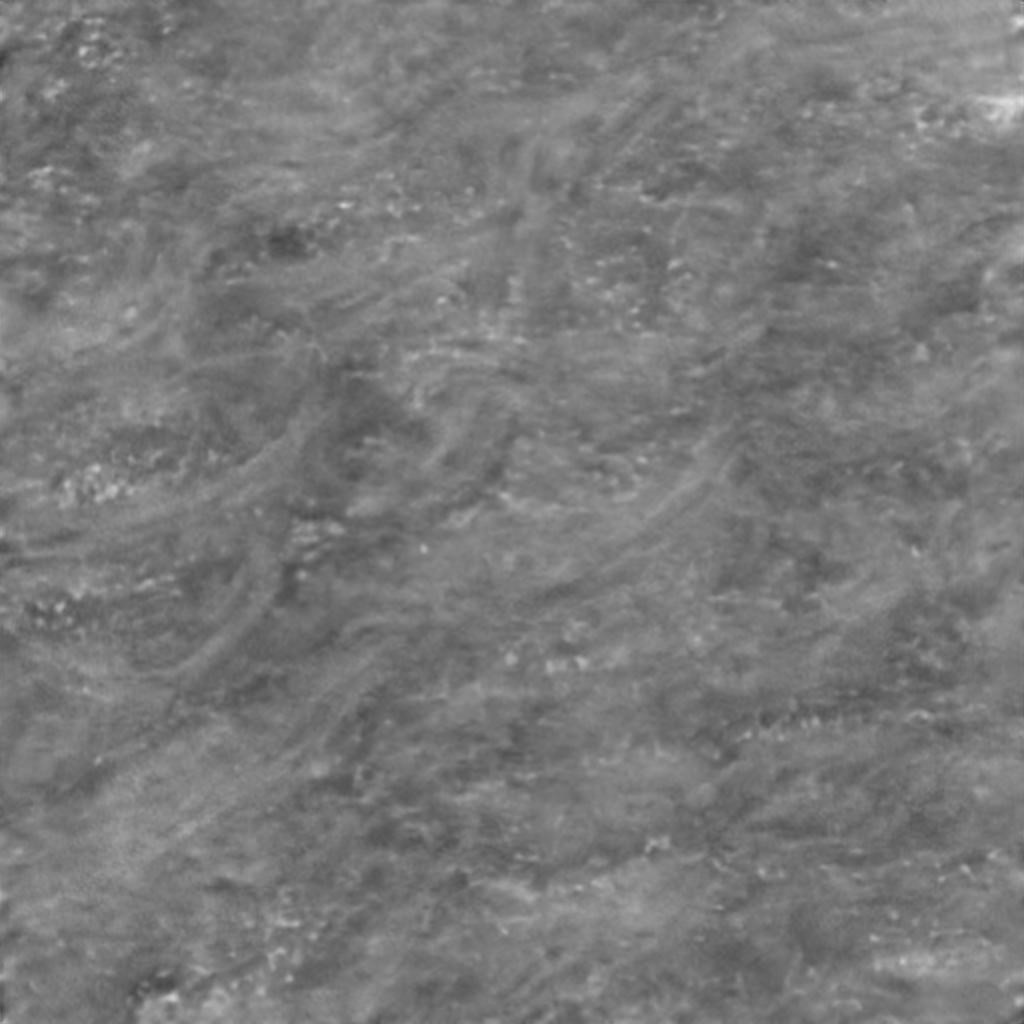}
    \vspace{-0.8\baselineskip}
\end{subfigure}
\begin{subfigure}[b]{0.11\textwidth}
    \includegraphics[width=\textwidth]{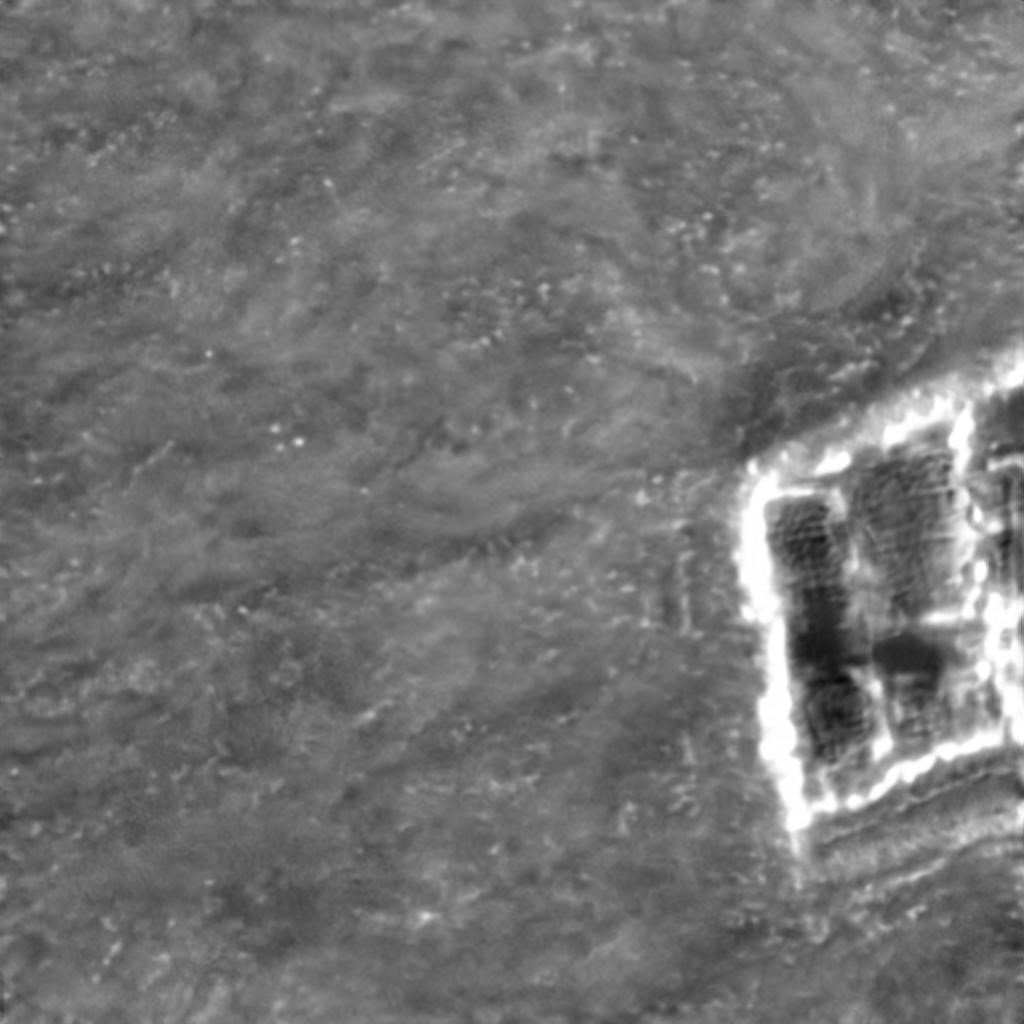}
    \vspace{-0.8\baselineskip}
\end{subfigure}
\begin{subfigure}[b]{0.11\textwidth}
    \includegraphics[width=\textwidth]{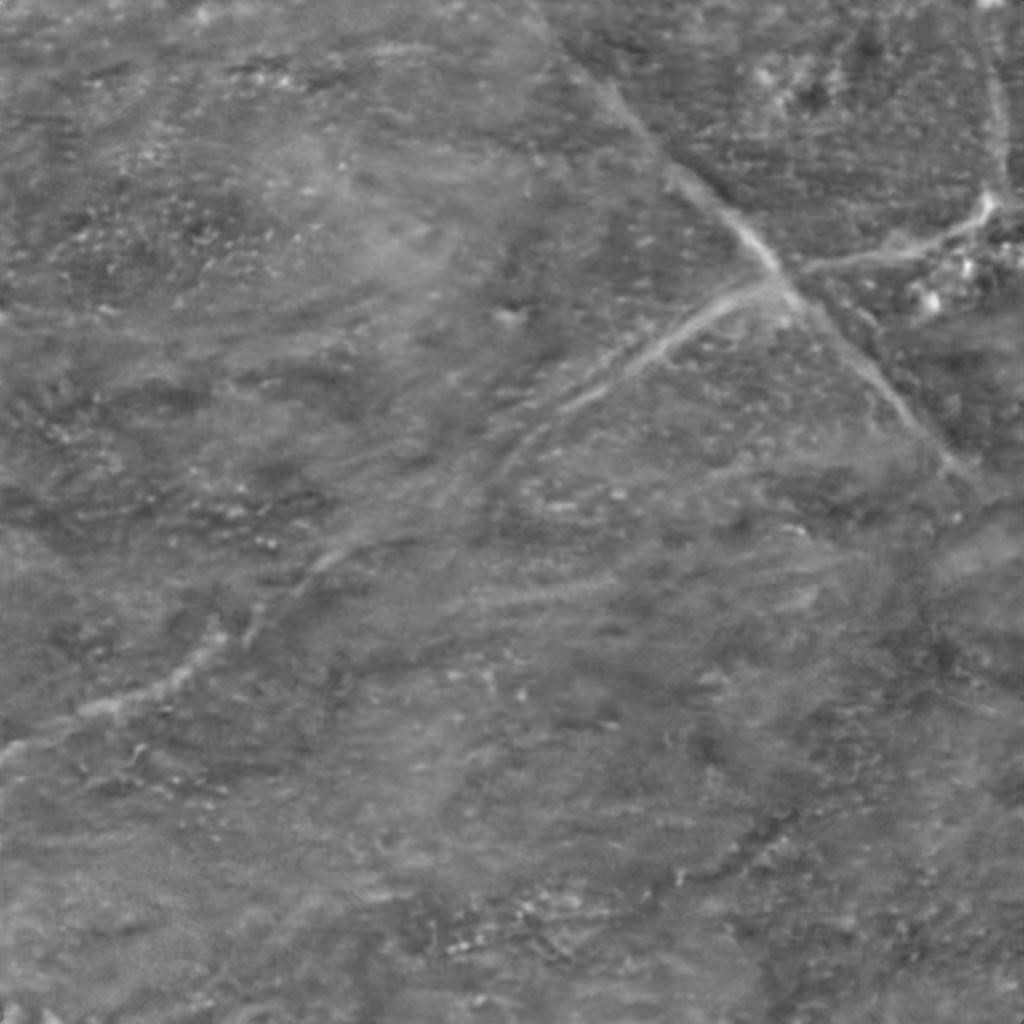}
    \vspace{-0.8\baselineskip}
\end{subfigure}
    \caption{Examples of satellite images, ground truth (GT) thermal images, and generated (Gen.) thermal images with or without CE in the test region.}
    \label{fig:generative}
\end{figure}

In Fig. \ref{fig:generative}, we show examples of satellite, ground truth, and generated thermal images with and without CE. We focus on the results with $\lambda_1=100.0$ since this setting as previously noticed gives us a better geo-localization performance compared to those with $\lambda_1=10.0$. The visualized results show that the thermal features of the farm and building portion are mostly clear and consistent in the generated results, while those of roads and deserts are distorted. For the results without CE, the thermal features of the desert portion are blurred and unrecognizable on the synthesized images, while the results with CE show detectable textures on that portion. We recognize that the existence of clear textures impacts the performance of the geo-localization model.

\begin{figure*}[ht]
    \centering
\smallskip
\smallskip
\rotatebox{90}{\scriptsize\hspace{2em}CE Thermal}
\begin{subfigure}[b]{0.11\textwidth}
    \includegraphics[width=\textwidth]{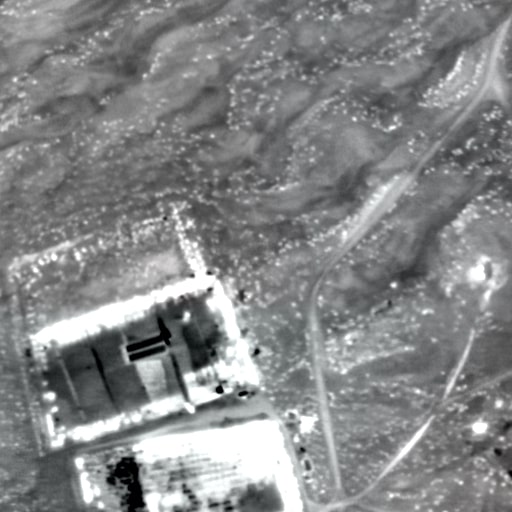}    \vspace{-0.8\baselineskip}
\end{subfigure}
\begin{subfigure}[b]{0.11\textwidth}
    \includegraphics[width=\textwidth]{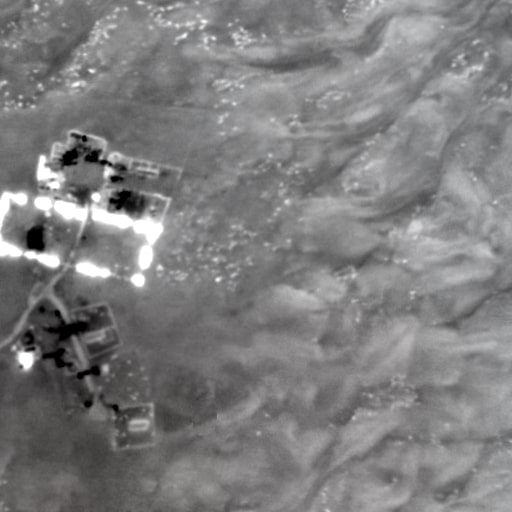}    \vspace{-0.8\baselineskip}
\end{subfigure}
\begin{subfigure}[b]{0.11\textwidth}
    \includegraphics[width=\textwidth]{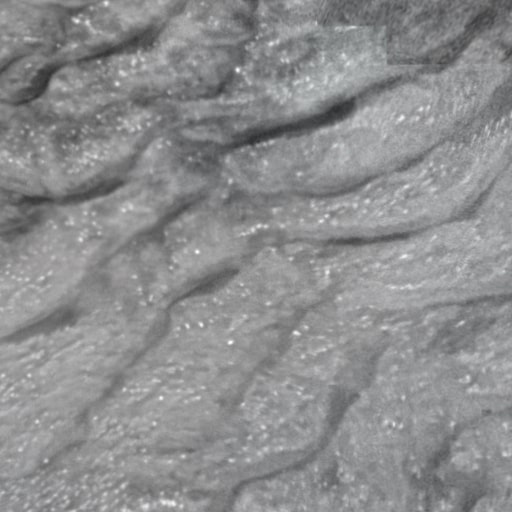}    \vspace{-0.8\baselineskip}
\end{subfigure}
\begin{subfigure}[b]{0.11\textwidth}
    \includegraphics[width=\textwidth]{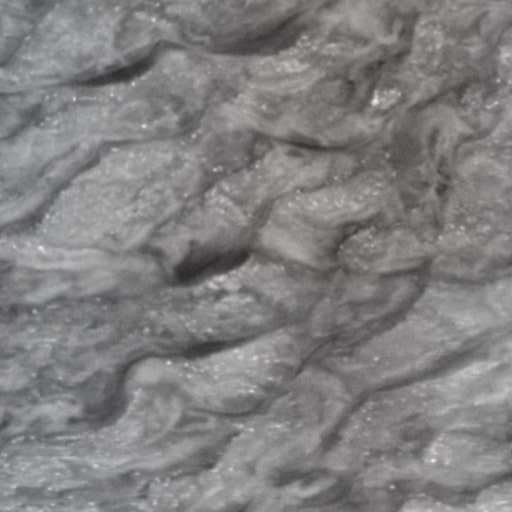}    \vspace{-0.8\baselineskip}
\end{subfigure}
\begin{subfigure}[b]{0.11\textwidth}
    \includegraphics[width=\textwidth]{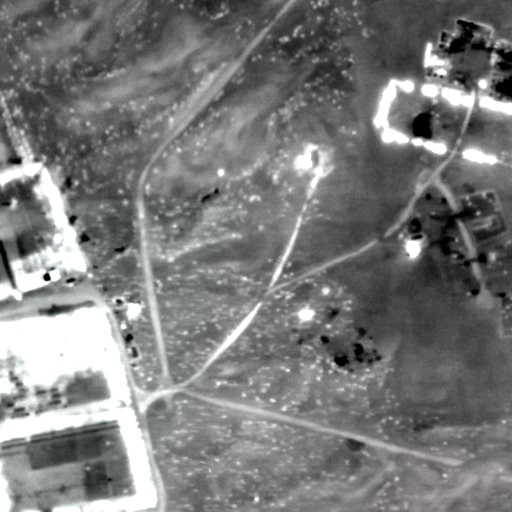}    \vspace{-0.8\baselineskip}
\end{subfigure}
\begin{subfigure}[b]{0.11\textwidth}
    \includegraphics[width=\textwidth]{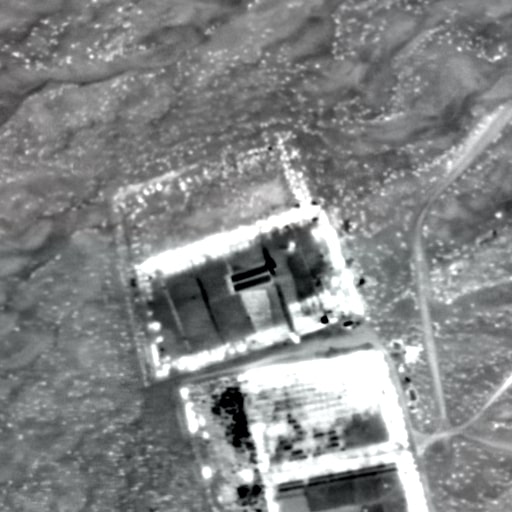}    \vspace{-0.8\baselineskip}
\end{subfigure}
\begin{subfigure}[b]{0.11\textwidth}
    \includegraphics[width=\textwidth]{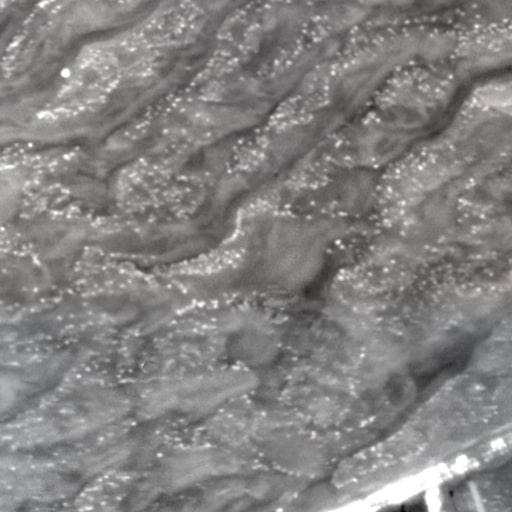}    \vspace{-0.8\baselineskip}
\end{subfigure}
\begin{subfigure}[b]{0.11\textwidth}
    \includegraphics[width=\textwidth]{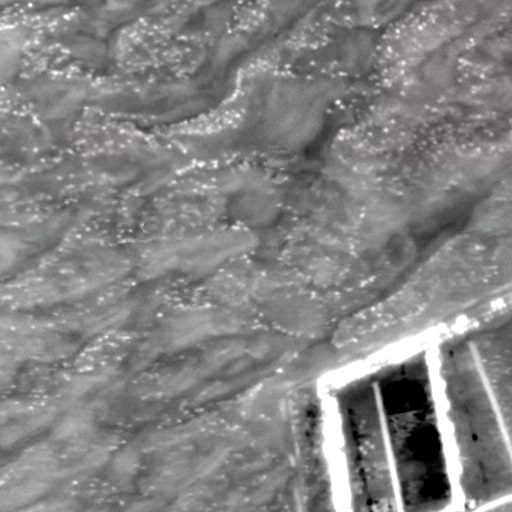}    \vspace{-0.8\baselineskip}
\end{subfigure}\\
\rotatebox{90}{\scriptsize\hspace{1em}Baseline Model}
\begin{subfigure}[b]{0.11\textwidth}
    \includegraphics[width=\textwidth]{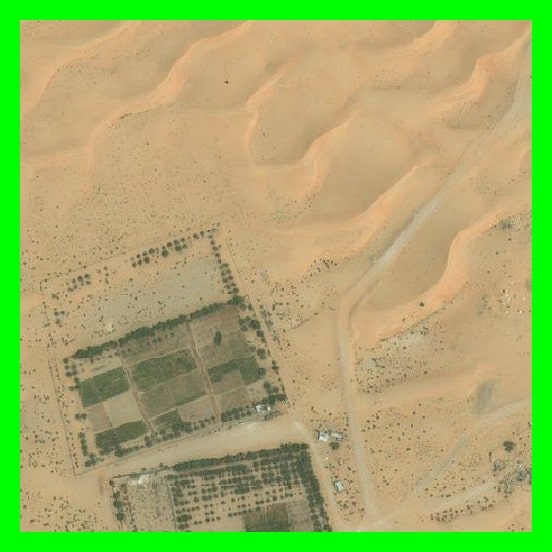}    \vspace{-0.8\baselineskip}
\end{subfigure}
\begin{subfigure}[b]{0.11\textwidth}
    \includegraphics[width=\textwidth]{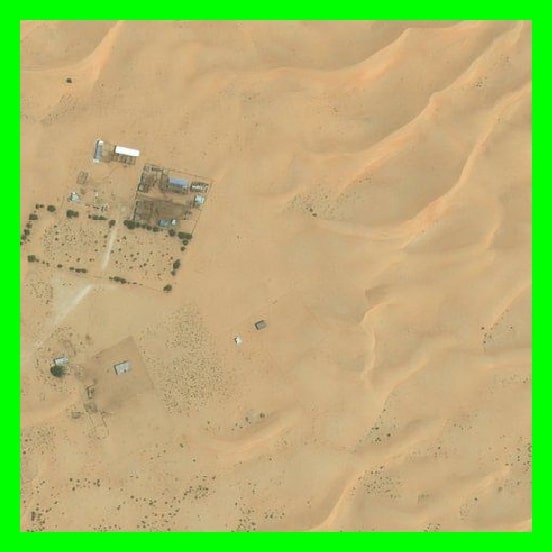}    \vspace{-0.8\baselineskip}
\end{subfigure}
\begin{subfigure}[b]{0.11\textwidth}
    \includegraphics[width=\textwidth]{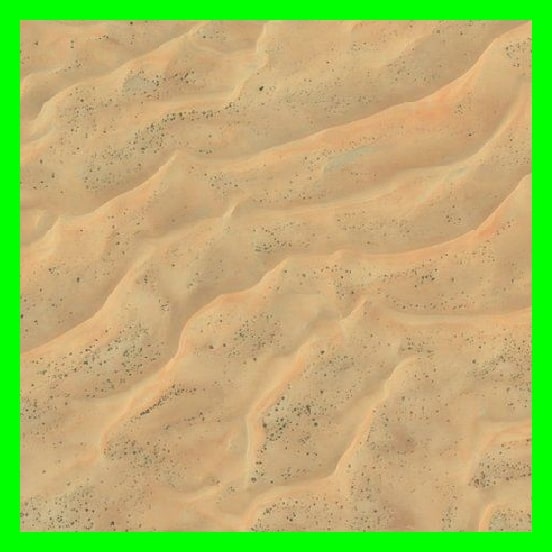}    \vspace{-0.8\baselineskip}
\end{subfigure}
\begin{subfigure}[b]{0.11\textwidth}
    \includegraphics[width=\textwidth]{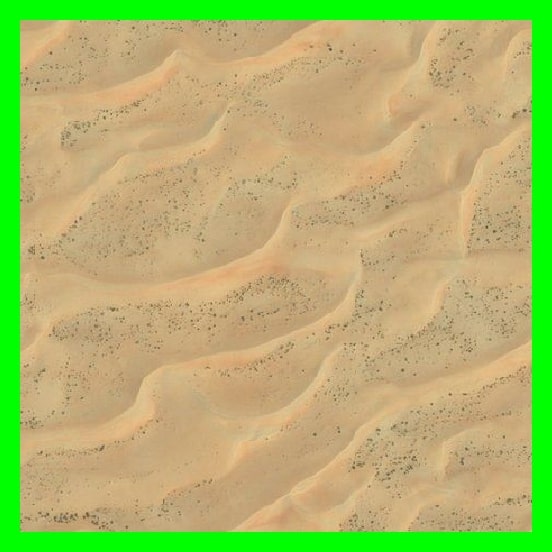}    \vspace{-0.8\baselineskip}
\end{subfigure}
\begin{subfigure}[b]{0.11\textwidth}
    \includegraphics[width=\textwidth]{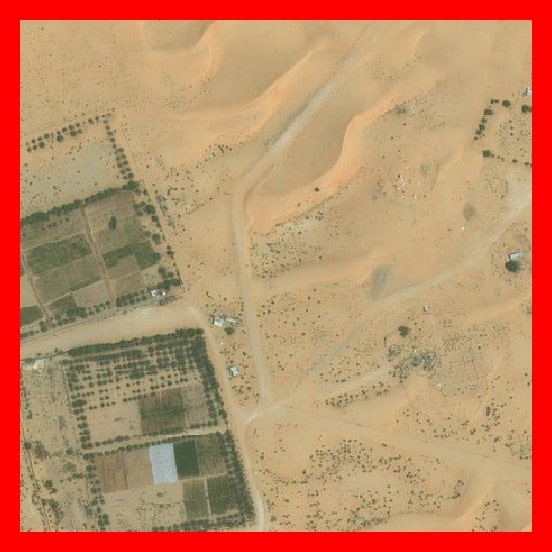}    \vspace{-0.8\baselineskip}
\end{subfigure}
\begin{subfigure}[b]{0.11\textwidth}
    \includegraphics[width=\textwidth]{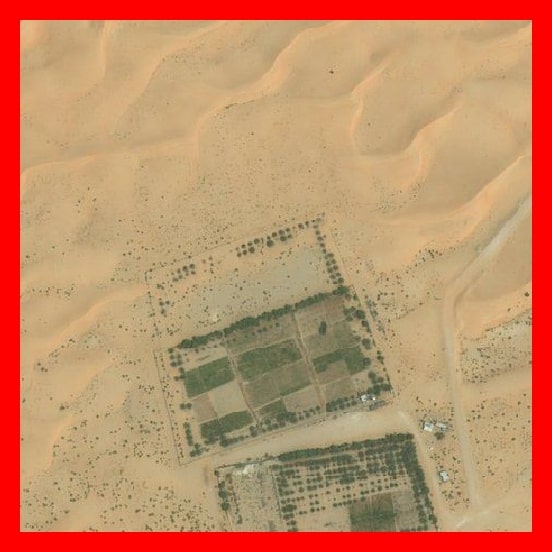}    \vspace{-0.8\baselineskip}
\end{subfigure}
\begin{subfigure}[b]{0.11\textwidth}
    \includegraphics[width=\textwidth]{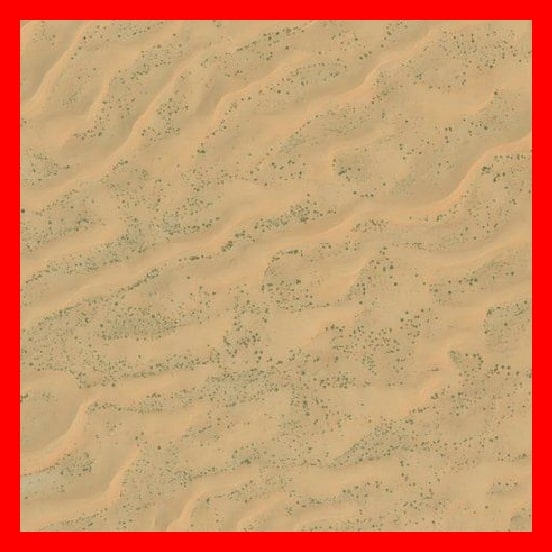}    \vspace{-0.8\baselineskip}
\end{subfigure}
\begin{subfigure}[b]{0.11\textwidth}
    \includegraphics[width=\textwidth]{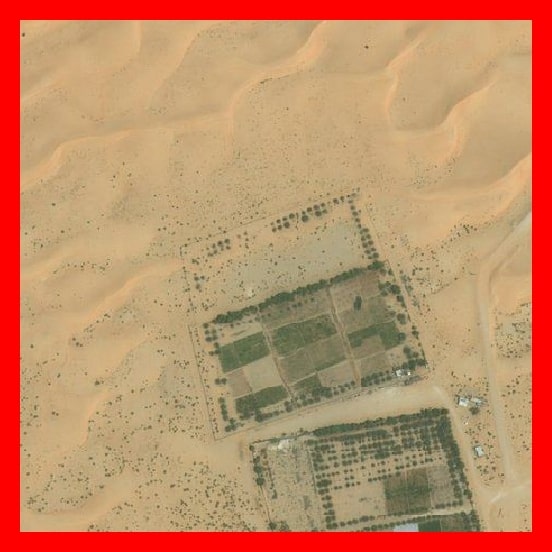}    \vspace{-0.8\baselineskip}
\end{subfigure}\\
\rotatebox{90}{\scriptsize\hspace{0.5em}Our Best Model}
\begin{subfigure}[b]{0.11\textwidth}
    \includegraphics[width=\textwidth]{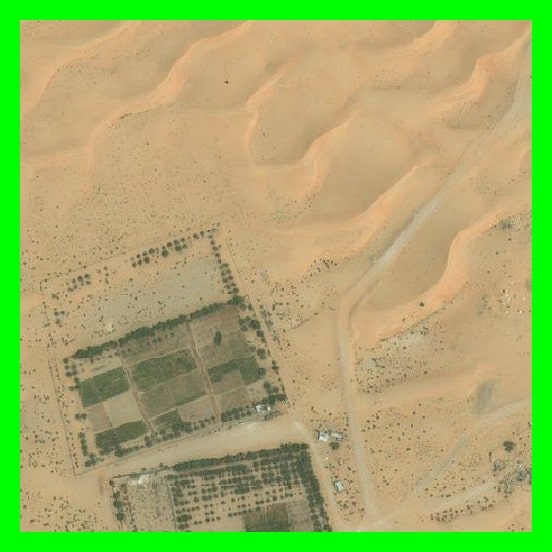}    
\end{subfigure}
\begin{subfigure}[b]{0.11\textwidth}
    \includegraphics[width=\textwidth]{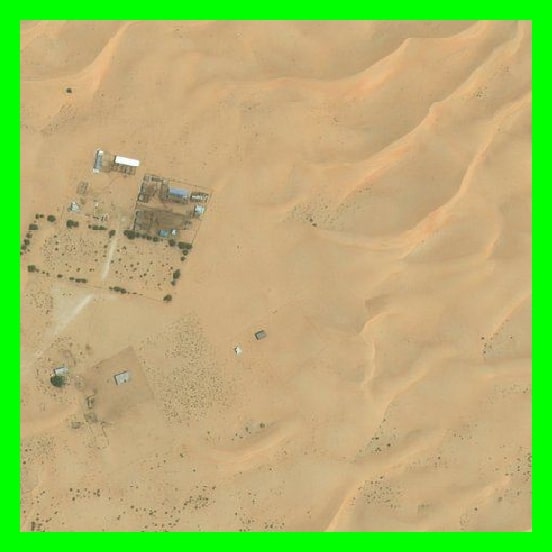}    
\end{subfigure}
\begin{subfigure}[b]{0.11\textwidth}
    \includegraphics[width=\textwidth]{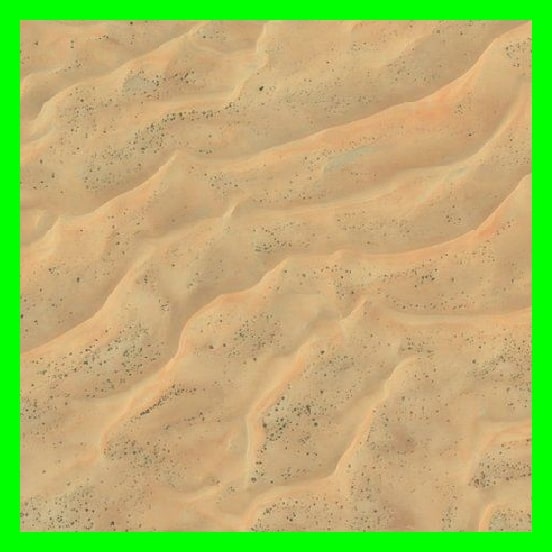}    
\end{subfigure}
\begin{subfigure}[b]{0.11\textwidth}
    \includegraphics[width=\textwidth]{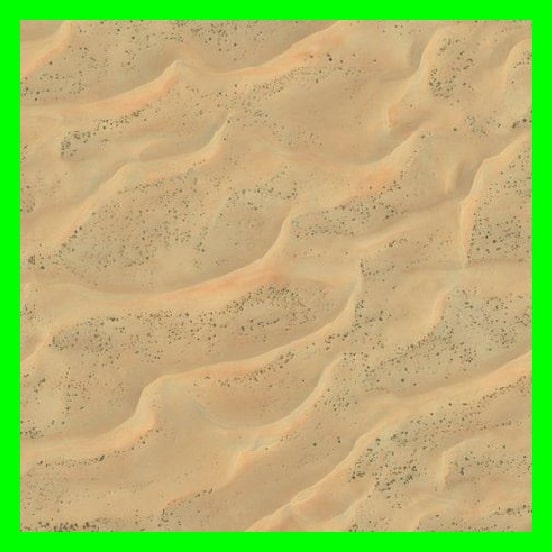}    
\end{subfigure}
\begin{subfigure}[b]{0.11\textwidth}
    \includegraphics[width=\textwidth]{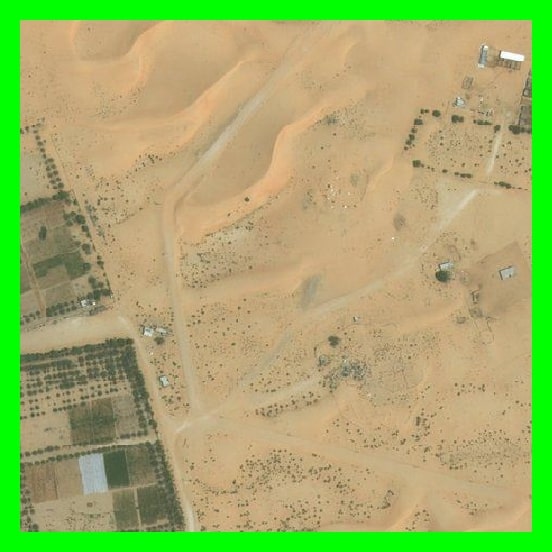}
\end{subfigure}
\begin{subfigure}[b]{0.11\textwidth}
    \includegraphics[width=\textwidth]{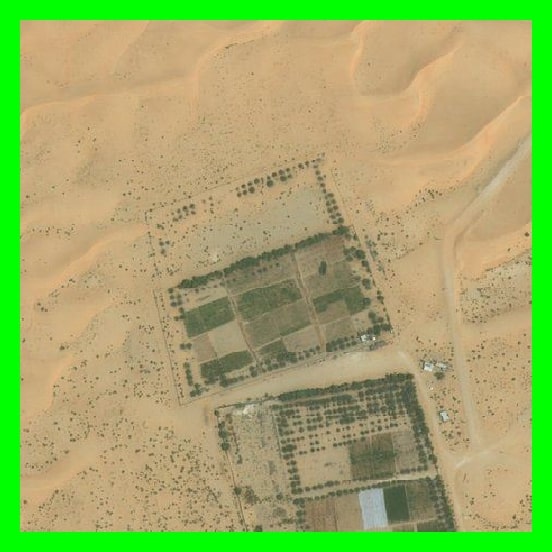}
\end{subfigure}
\begin{subfigure}[b]{0.11\textwidth}
    \includegraphics[width=\textwidth]{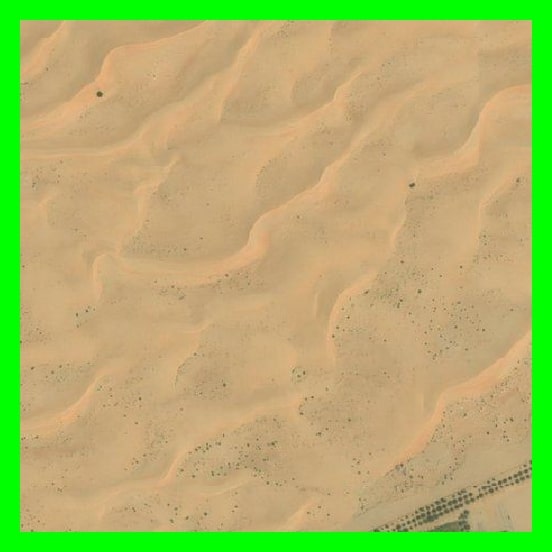}
\end{subfigure}
\begin{subfigure}[b]{0.11\textwidth}
    \includegraphics[width=\textwidth]{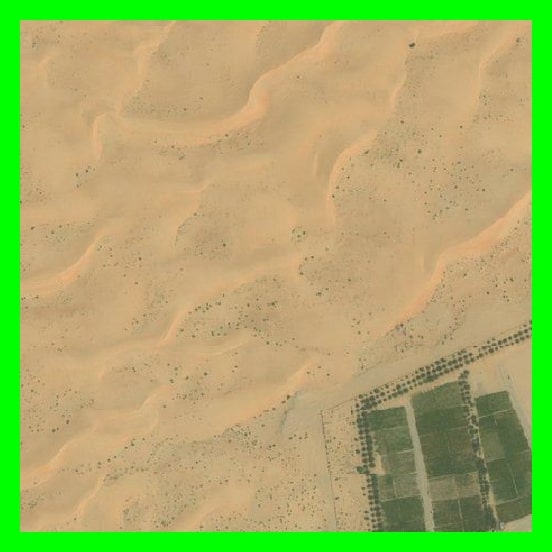}
\end{subfigure}
\caption{Examples of ground truth thermal images with CE, correct and failed Top-1 retrieved satellite images in the test region. The $1^\text{st}$ row is the input thermal images with CE. The $2^\text{nd}$ (Baseline Model) and $3^\text{rd}$ (Our Best Model) rows are correct and failed cases comparing different settings. The correct ($L_2$ distance error $\leq 35$\,m) and failed ($L_2$ distance error $\geq 50$\,m) ones are bounded by green and red colors, respectively. The $5^\text{th}$ and $6^\text{th}$ columns of the $2^\text{nd}$ row show offset errors. The $7^\text{th}$ and $8^\text{th}$ columns show localization failures.}
    \label{fig:geo}
    \vspace{-10pt}
\end{figure*}

\subsection{Visualized Geo-localization Results}\label{sec:geo}
We compare the visualized SGM results of the Baseline Model and Our Best Model in Fig. \ref{fig:geo}. Our framework can retrieve accurate results on both farm regions and desert regions with the indistinct self-similar thermal feature, as shown in the examples. However, the baseline setting shows more errors than our best setting. We identify two types of failure cases: Offset error and localization failure. Offset error results in the substantial offset ($\geq 50$\,m) of the retrieved satellite image from the true one. The $5^\text{th}$ and $6^\text{th}$ columns of the $2^\text{nd}$ row in Fig. \ref{fig:geo} show the matching result offsetting from the true position. Localization failure results in completely wrong retrieved results, which may cause a localization system failure. The $7^\text{th}$ and $8^\text{th}$ columns of the $2^\text{nd}$ row in Fig. \ref{fig:geo} randomly match the wrong satellite images. We recognize that the main reason for the above errors is the self-similar low-contrast thermal features in the desert. The Baseline Model suffers from limited paired data and self-similar low-contrast thermal features and makes more offset errors and localization failures, while Our Best Model mitigates the above problems. Visually, Our Best Model is demonstrated to outperform the Baseline Model. 

\section{Conclusions}~\label{sec:conclusions}
In this work, we presented a novel thermal geo-localization framework using satellite imagery for high-altitude long-range UAV flight geo-localization. The baseline model suffers from the lack of paired satellite and thermal images. To address this, our framework combine adversarial-based domain adaptation techniques with a generative model. The proposed approach exploits numerous unpaired satellite images to mitigate the limitation of paired data. The experiments show that our framework delivers notable improvement in geo-localization performances even in an test region with low-contrast self-similar thermal features. Future work will focus on further reducing the output dimension for computation efficiency and end-to-end training for TGM and SGM for real-time deployment.
\bibliographystyle{IEEEtran} 
\bibliography{mybib}

\begin{thebibliography}{10}
\providecommand{\url}[1]{#1}
\csname url@samestyle\endcsname
\providecommand{\newblock}{\relax}
\providecommand{\bibinfo}[2]{#2}
\providecommand{\BIBentrySTDinterwordspacing}{\spaceskip=0pt\relax}
\providecommand{\BIBentryALTinterwordstretchfactor}{4}
\providecommand{\BIBentryALTinterwordspacing}{\spaceskip=\fontdimen2\font plus
\BIBentryALTinterwordstretchfactor\fontdimen3\font minus
  \fontdimen4\font\relax}
\providecommand{\BIBforeignlanguage}[2]{{%
\expandafter\ifx\csname l@#1\endcsname\relax
\typeout{** WARNING: IEEEtran.bst: No hyphenation pattern has been}%
\typeout{** loaded for the language `#1'. Using the pattern for}%
\typeout{** the default language instead.}%
\else
\language=\csname l@#1\endcsname
\fi
#2}}
\providecommand{\BIBdecl}{\relax}
\BIBdecl

\bibitem{review_avl}
A.~Couturier and M.~A. Akhloufi, ``A review on absolute visual localization for
  uav,'' \emph{Robotics and Autonomous Systems}, vol. 135, p. 103666, 2021.

\bibitem{driftingerror}
S.~Weiss, M.~W. Achtelik, S.~Lynen, M.~C. Achtelik, L.~Kneip, M.~Chli, and
  R.~Siegwart, ``Monocular vision for long-term micro aerial vehicle state
  estimation: A compendium,'' \emph{Journal of Field Robotics}, vol.~30, no.~5,
  pp. 803--831, 2013.

\bibitem{directalign}
G.~J.~V. Dalen, D.~P. Magree, and E.~N. Johnson, ``Absolute localization using
  image alignment and particle filtering,'' in \emph{AIAA Guidance, Navigation,
  and Control Conference}, 2016.

\bibitem{directalign2}
A.~Yol, B.~Delabarre, A.~Dame, J.-E. Dartois, and E.~Marchand, ``Vision-based
  absolute localization for unmanned aerial vehicles,'' in \emph{IEEE/RSJ
  International Conference on Intelligent Robots and Systems}, 2014, pp.
  3429--3434.

\bibitem{directalign3}
B.~Patel, T.~D. Barfoot, and A.~P. Schoellig, ``Visual localization with google
  earth images for robust global pose estimation of uavs,'' in \emph{IEEE
  International Conference on Robotics and Automation (ICRA)}, 2020, pp.
  6491--6497.

\bibitem{imgregistration}
M.~Shan, F.~Wang, F.~Lin, Z.~Gao, Y.~Z. Tang, and B.~M. Chen, ``Google map
  aided visual navigation for uavs in gps-denied environment,'' in \emph{IEEE
  International Conference on Robotics and Biomimetics (ROBIO)}, 2015, pp.
  114--119.

\bibitem{imgregistration2}
M.~Mantelli, D.~Pittol, R.~Neuland, A.~Ribacki, R.~Maffei, V.~Jorge,
  E.~Prestes, and M.~Kolberg, ``A novel measurement model based on abbrief for
  global localization of a uav over satellite images,'' \emph{Robotics and
  Autonomous Systems}, vol. 112, pp. 304--319, 2019.

\bibitem{Berton_CVPR_2022_benchmark}
G.~Berton, R.~Mereu, G.~Trivigno, C.~Masone, G.~Csurka, T.~Sattler, and
  B.~Caputo, ``Deep visual geo-localization benchmark,'' in \emph{IEEE/CVF
  Conference on Computer Vision and Pattern Recognition (CVPR)}, 2022.

\bibitem{netvlad}
R.~Arandjelović, P.~Gronat, A.~Torii, T.~Pajdla, and J.~Sivic, ``Netvlad: Cnn
  architecture for weakly supervised place recognition,'' \emph{IEEE
  Transactions on Pattern Analysis and Machine Intelligence}, vol.~40, no.~6,
  pp. 1437--1451, 2018.

\bibitem{VG1}
Y.~Ge, H.~Wang, F.~Zhu, R.~Zhao, and H.~Li, ``Self-supervising fine-grained
  region similarities for large-scale image localization,'' in \emph{Computer
  Vision -- ECCV 2020}, A.~Vedaldi, H.~Bischof, T.~Brox, and J.-M. Frahm,
  Eds.\hskip 1em plus 0.5em minus 0.4em\relax Cham: Springer International
  Publishing, 2020, pp. 369--386.

\bibitem{patch-netvlad}
S.~Hausler, S.~Garg, M.~Xu, M.~Milford, and T.~Fischer, ``Patch-netvlad:
  Multi-scale fusion of locally-global descriptors for place recognition,'' in
  \emph{IEEE/CVF Conference on Computer Vision and Pattern Recognition (CVPR)},
  2021, pp. 14\,136--14\,147.

\bibitem{VG3}
G.~Berton, C.~Masone, V.~Paolicelli, and B.~Caputo, ``Viewpoint invariant dense
  matching for visual geolocalization,'' in \emph{IEEE/CVF International
  Conference on Computer Vision (ICCV)}, 2021, pp. 12\,149--12\,158.

\bibitem{lecun2015deep}
Y.~LeCun, Y.~Bengio, and G.~Hinton, ``Deep learning,'' \emph{nature}, vol. 521,
  no. 7553, pp. 436--444, 2015.

\bibitem{bow}
Sivic and Zisserman, ``Video google: a text retrieval approach to object
  matching in videos,'' in \emph{Proceedings Ninth IEEE International
  Conference on Computer Vision}, 2003, pp. 1470--1477.

\bibitem{vlad}
H.~Jégou, M.~Douze, C.~Schmid, and P.~Pérez, ``Aggregating local descriptors
  into a compact image representation,'' in \emph{IEEE/CVF Conference on
  Computer Vision and Pattern Recognition (CVPR)}, 2010, pp. 3304--3311.

\bibitem{SIFT}
D.~G. Lowe, ``Distinctive image features from scale-invariant keypoints,''
  \emph{International journal of computer vision}, vol.~60, no.~2, pp. 91--110,
  2004.

\bibitem{msls}
F.~Warburg, S.~Hauberg, M.~Lopez-Antequera, P.~Gargallo, Y.~Kuang, and
  J.~Civera, ``Mapillary street-level sequences: A dataset for lifelong place
  recognition,'' in \emph{IEEE/CVF Conference on Computer Vision and Pattern
  Recognition (CVPR)}, 2020.

\bibitem{voplusvg}
A.~Shetty and G.~X. Gao, ``Uav pose estimation using cross-view geolocalization
  with satellite imagery,'' in \emph{IEEE International Conference on Robotics
  and Automation (ICRA)}, 2019, pp. 1827--1833.

\bibitem{style}
X.~Tian, J.~Shao, D.~Ouyang, and H.~T. Shen, ``Uav-satellite view synthesis for
  cross-view geo-localization,'' \emph{IEEE Transactions on Circuits and
  Systems for Video Technology}, vol.~32, no.~7, pp. 4804--4815, 2022.

\bibitem{VGUAV3}
S.~Chen, X.~Wu, M.~W. Mueller, and K.~Sreenath, ``Real-time geo-localization
  using satellite imagery and topography for unmanned aerial vehicles,'' in
  \emph{IEEE/RSJ International Conference on Intelligent Robots and Systems
  (IROS)}, 2021, pp. 2275--2281.

\bibitem{VGUAV4}
M.~Bianchi and T.~D. Barfoot, ``Uav localization using autoencoded satellite
  images,'' \emph{IEEE Robotics and Automation Letters}, vol.~6, no.~2, pp.
  1761--1768, 2021.

\bibitem{VGUAV5}
H.~Goforth and S.~Lucey, ``Gps-denied uav localization using pre-existing
  satellite imagery,'' in \emph{IEEE International Conference on Robotics and
  Automation (ICRA)}, 2019, pp. 2974--2980.

\bibitem{cgan}
M.~Mirza and S.~Osindero, ``Conditional generative adversarial nets,''
  \emph{arXiv preprint arXiv:1411.1784}, 2014.

\bibitem{thermaluav1}
T.~Mouats, N.~Aouf, L.~Chermak, and M.~A. Richardson, ``Thermal stereo odometry
  for uavs,'' \emph{IEEE Sensors Journal}, vol.~15, no.~11, pp. 6335--6347,
  2015.

\bibitem{thermaluav2}
S.~Khattak, C.~Papachristos, and K.~Alexis, ``Keyframe-based direct
  thermal--inertial odometry,'' in \emph{International Conference on Robotics
  and Automation (ICRA)}, 2019, pp. 3563--3569.

\bibitem{thermaluav3}
J.~Delaune, R.~Hewitt, L.~Lytle, C.~Sorice, R.~Thakker, and L.~Matthies,
  ``Thermal-inertial odometry for autonomous flight throughout the night,'' in
  \emph{IEEE/RSJ International Conference on Intelligent Robots and Systems
  (IROS)}, 2019, pp. 1122--1128.

\bibitem{thermaluav4}
V.~Polizzi, R.~Hewitt, J.~Hidalgo-Carrió, J.~Delaune, and D.~Scaramuzza,
  ``Data-efficient collaborative decentralized thermal-inertial odometry,''
  \emph{IEEE Robotics and Automation Letters}, vol.~7, no.~4, pp.
  10\,681--10\,688, 2022.

\bibitem{DANN}
Y.~Ganin and V.~Lempitsky, ``Unsupervised domain adaptation by
  backpropagation,'' in \emph{International Conference on Machine Learning
  (ICML)}, 2015, pp. 1180--1189.

\bibitem{cycada}
J.~Hoffman, E.~Tzeng, T.~Park, J.-Y. Zhu, P.~Isola, K.~Saenko, A.~Efros, and
  T.~Darrell, ``{C}y{CADA}: Cycle-consistent adversarial domain adaptation,''
  in \emph{International Conference on Machine Learning (ICML)}, 2018, pp.
  1989--1998.

\bibitem{DA1}
E.~Tzeng, J.~Hoffman, K.~Saenko, and T.~Darrell, ``Adversarial discriminative
  domain adaptation,'' in \emph{Proceedings of the IEEE Conference on Computer
  Vision and Pattern Recognition (CVPR)}, 2017.

\bibitem{thermalDA1}
L.~Gan, C.~Lee, and S.-J. Chung, ``Unsupervised rgb-to-thermal domain
  adaptation via multi-domain attention network,'' \emph{arXiv preprint
  arXiv:2210.04367}, 2022.

\bibitem{thermalDA2}
Y.-H. Kim, U.~Shin, J.~Park, and I.~S. Kweon, ``Ms-uda: Multi-spectral
  unsupervised domain adaptation for thermal image semantic segmentation,''
  \emph{IEEE Robotics and Automation Letters}, vol.~6, no.~4, pp. 6497--6504,
  2021.

\bibitem{thermalDA3}
I.~B. Akkaya, F.~Altinel, and U.~Halici, ``Self-training guided adversarial
  domain adaptation for thermal imagery,'' in \emph{IEEE/CVF Conference on
  Computer Vision and Pattern Recognition Workshops (CVPRW)}, 2021, pp.
  4317--4326.

\bibitem{thermalDA4}
V.~Vs, D.~Poster, S.~You, S.~Hu, and V.~M. Patel, ``Meta-uda: Unsupervised
  domain adaptive thermal object detection using meta-learning,'' in
  \emph{IEEE/CVF Winter Conference on Applications of Computer Vision (WACV)},
  2022, pp. 3697--3706.

\bibitem{CycleGAN2017}
J.-Y. Zhu, T.~Park, P.~Isola, and A.~A. Efros, ``Unpaired image-to-image
  translation using cycle-consistent adversarial networks,'' in \emph{IEEE
  International Conference on Computer Vision (ICCV)}, 2017.

\bibitem{thermalpix2pix}
X.~Qian, M.~Zhang, and F.~Zhang, ``Sparse gans for thermal infrared image
  generation from optical image,'' \emph{IEEE Access}, vol.~8, pp.
  180\,124--180\,132, 2020.

\bibitem{pix2pix}
P.~Isola, J.-Y. Zhu, T.~Zhou, and A.~A. Efros, ``Image-to-image translation
  with conditional adversarial networks,'' in \emph{IEEE/CVF Conference on
  Computer Vision and Pattern Recognition (CVPR)}, 2017, pp. 1125--1134.

\bibitem{lsgan}
X.~Mao, Q.~Li, H.~Xie, R.~Y. Lau, Z.~Wang, and S.~Paul~Smolley, ``Least squares
  generative adversarial networks,'' in \emph{Proceedings of the IEEE
  International Conference on Computer Vision (ICCV)}, 2017.

\bibitem{unet}
O.~Ronneberger, P.~Fischer, and T.~Brox, ``U-net: Convolutional networks for
  biomedical image segmentation,'' in \emph{Medical Image Computing and
  Computer-Assisted Intervention (MICCAI)}, 2015, pp. 234--241.

\bibitem{adam}
D.~P. Kingma and J.~Ba, ``Adam: A method for stochastic optimization,'' in
  \emph{International Conference on Learning Representations (ICLR)}, 2015.

\bibitem{resnet}
K.~He, X.~Zhang, S.~Ren, and J.~Sun, ``Deep residual learning for image
  recognition,'' in \emph{IEEE International Conference on Computer Vision and
  Pattern Recognition (CVPR)}, 2016, pp. 770--778.

\bibitem{NEURIPS2019_9015}
A.~Paszke, S.~Gross, F.~Massa, A.~Lerer, J.~Bradbury, G.~Chanan, T.~Killeen,
  Z.~Lin, N.~Gimelshein, L.~Antiga, A.~Desmaison, A.~Kopf, E.~Yang, Z.~DeVito,
  M.~Raison, A.~Tejani, S.~Chilamkurthy, B.~Steiner, L.~Fang, J.~Bai, and
  S.~Chintala, ``Pytorch: An imperative style, high-performance deep learning
  library,'' in \emph{Conference on Neural Information Processing Systems
  (NeurIPS)}, 2019, pp. 8024--8035.

\end{thebibliography}

\end{document}